\documentclass{article}

 \usepackage[final,nonatbib]{neurips_2022}

\usepackage{bbm}
\usepackage{graphicx}
\usepackage{amsmath}
\usepackage{enumitem}

\usepackage[utf8]{inputenc} 
\usepackage[T1]{fontenc}    
\usepackage[ hypertexnames=false, bookmarksnumbered = True, colorlinks=true, allcolors=blue]{hyperref} 

\usepackage{url}            
\usepackage{booktabs}       
\usepackage{amsfonts}       
\usepackage{nicefrac}       
\usepackage{microtype}      
\usepackage{xcolor}         

\usepackage{cleveref} 
\crefname{assumption}{Assumption}{Assumptions}
\crefname{definition}{Definition}{Definitions}
\crefname{lemma}{Lemma}{Lemmas}
\crefname{theorem}{Theorem}{Theorems}
\crefname{corollary}{Corollary}{Corollaries}
\crefname{proposition}{Proposition}{Propositions}
\crefname{claim}{Claim}{Claims}
\crefname{subclaim}{Subclaim}{Subclaims}
\crefname{procedure}{Procedure}{Procedures}
\crefname{algorithm}{Algorithm}{Algorithms}
\crefname{example}{Example}{Examples}
\crefname{figure}{Figure}{Figures}
\crefname{section}{Section}{Sections}
\crefname{appendix}{Appendix}{Appendices}
\crefname{table}{Table}{Tables}
\crefname{equation}{}{}
\crefname{appsec}{Appendix}{Appendices}
\crefname{fact}{Fact}{Facts}

\newtheorem{theorem}{Theorem}

\newtheorem{lemma}{Lemma}

\title{On A Mallows-type Model For (Ranked) Choices}

\author{%
	Yifan Feng
	\\
	Department of Analytics and Operations\\
	NUS Business School\\
	\texttt{yifan.feng@nus.edu.sg}
	\And
	Yuxuan Tang
	\\
	Institute of Operations Research and Analytics\\
	National University of Singapore\\
	\texttt{yuxuan.tang@u.nus.edu}
}

\newenvironment{proofref}[1]{\noindent {\bf Proof of {#1}.\/}}{$\square$\vskip 0.01in}

\newcommand{\ssum}{\sum\nolimits}
\newcommand{\sprod}{\prod\nolimits}
\newcommand{\tpi}{\tilde{\pi}}
\newcommand{\I}[1]{\mathbb{I} \{#1\}}
\newcommand{\II}[1]{\mathbf{I} (#1)}
\newcommand{\calS}{\mathcal{S}}
\newcommand{\calC}{\mathcal{C}}

\usepackage{tikz}

\begin{document}

\maketitle

\begin{abstract}
We consider a preference learning setting where every participant chooses an ordered list of $k$ most preferred items among a displayed set of candidates. (The set can be different for every participant.) We identify a distance-based ranking model for the population's preferences and their (ranked) choice behavior. The ranking model resembles the Mallows model but uses a new distance function called Reverse Major Index (RMJ). We find that despite the need to sum over all permutations, the RMJ-based ranking distribution aggregates into (ranked) choice probabilities with simple closed-form expression. We develop effective methods to estimate the model parameters and showcase their generalization power using real data, especially when there is a limited variety of display sets.
\end{abstract}

\section{Introduction}

How to aggregate the population's preferences from their (ranked) choices out of different choice sets? This question is of interest to many communities, such as economics, business, and computer science. A concrete setting is a platform that wishes to learn customer preferences over a universe of $ n $ product prototypes. The platform is able to display different subsets of versions to different customers, who then provide feedback in the form of a top-$ k $ ranked list of their most preferred items within the subset they see (hereafter referred to as ``ranked choices").\footnote{If $ k = 1 $, each participant is asked to choose the most preferred candidate. That effectively reduces to a (single) choice, which is a feedback structure extensively studied in the choice modeling literature.}

The population's (ranked) choice behavior could be summarized as a \textit{(ranked) choice model} $ \Pr(\pi_k|S) $, which specifies the probability that a randomly drawn participant choosing a top-$ k $ list $ \pi_k $ from the display set $ S $. An economically rationalizable and yet very general way to model (ranked) choices is to use probabilistic \textit{ranking models}. That is, given a probability distribution $ \lambda $ over preference rankings, $ \Pr(\pi_k|S) $ equals the probability that a randomly drawn participant would place $ \pi_k $ as the top-$ k $ positions among the items in $ S $. 

A popular family of ranking models is \textit{distance-based}, which is the conceptual analog of Gaussian distribution for scalars; see \cite{fligner1986distance}. A distance-based ranking model is specified by a modal (central) ranking $ \pi^\ast $ and a dispersion parameter $ q \in (0,1) $. Given a ranking $ \pi $, its probability of being sampled is proportional to $ q^{d(\pi^\ast, \pi)} $. Here $ d(\pi_1, \pi_2) $ is a \textit{distance function} that describes the discrepancy of ranking $ \pi_1 $ from $ \pi_2 $, and different distance functions lead to different models. The most popular distance-based ranking model is the \textit{Mallows model} (\cite{mallows1957non}), which uses the Kendall-Tau distance as its distance function. It has been studied extensively in the literature regarding topics such as sampling, estimation from sampled (partial) rankings, and learning in a Mallows mixture setting; see \cite{chierichetti2018mallows} and references therein. Due to its popularity and the fact that every distance-based ranking model only differs from the Mallows by distance function, we will also refer to a distance-based ranking model as a \textit{Mallows-type model} interchangeably in the sequel.

Mallows-type models could be used as ``kernels" to ``smooth out" the distribution over rankings. As such, they could help mitigate the overfitting issues of generic (i.e., nonparametric) ranking models, which are typically overparameterized. Yet, Mallows models can be highly expressive in a mixture setting: A mixture of Mallows-type models can approximate \textit{any} probability distribution over rankings by an arbitrary precision (as $ q $ tends to zero and the number of clusters tends to infinity). Therefore, the Mallows-type model family is a helpful tool to balance capturing (complex) preference heterogeneity across individuals and regularizing the ranking distribution for better out-of-sample predictions. A representative work is by Antoine et al. \cite{Antoine2021}, who use the Mallows model to aggregate customer preferences from their choices (i.e., $k=1$). 

Despite the theoretical elegance, the main challenge in applying Mallows-type ranking models to (ranked) choice modeling is analytical and computational tractability. More specifically, $ \Pr(\pi_k|S) $ is calculated from summation over all rankings subject to nontrivial conditions. Therefore, even if a Mallows-type ranking model has a simple structure for $ \lambda $, the resulting choice probabilities $ \{\Pr(\pi_k | S)\} $ can be difficult to obtain even when $ k = 1 $. The state-of-art results are obtained by Antoine et al. \cite{Antoine2021}, who develop polynomial-time numerical algorithms to compute $ \{\Pr(\pi_k | S)\} $ for $ k =1 $ under the Mallows model. The \textit{estimation} problem is even more difficult, which involves finding the central ranking $ \pi^\ast $ and dispersion parameters $ q $ that best explain the (ranked) choice data. To the best of our knowledge, no effective methods to estimate \textit{any} Mallows-type ranking model from (ranked) choice data are known. (Perhaps the best method to date is again by Antoine et al. \cite{Antoine2021}, who use a ``Mallows smoothing" heuristic to conduct the estimation when $k=1$, which we will discuss later.) 

\textbf{Summary of results and contributions.} This paper identifies and studies a new distance-based (i.e., Mallows-type) ranking model. It is the same as the Mallows model except that it builds on a new distance function (i.e., smoothing kernel), which we call \textit{reverse major index (RMJ)}. Unlike the Mallows model's Kendall-Tau distance (which weighs all pairwise disagreements equally), RMJ puts more weight on top-position deviations.

The RMJ-based ranking model is a small conceptual deviation from the Mallows model (and, in particular, enjoys the desired properties mentioned above, such as rationalizability, smoothing, and expressive power); see \cref{sec:RMI model}. 
However, this twist brings a significant advantage in both analytical and computational simplicity. Specifically,  we solve a list of problems under the RMJ-based ranking distribution. That includes:
\begin{itemize}
	\item \textit{Characterizing (ranked) choice probabilities}: Given $ k \geq 1 $, calculating $ \{\Pr(\pi_k | S)\} $;
	\item \textit{Sampling}: Given $ k \geq 1 $, efficiently sampling a top-$ k $ list $ \pi_k $;
	\item \textit{Parameter learning}: Estimating the central ranking $ \pi^\ast $ and dispersion parameter $ q $ from the (ranked) choice data through a maximum likelihood estimator (MLE) formulation ;
	\item \textit{Learning in a mixture setting}: Assuming that there are multiple clusters of participant preferences, learning the central ranking and dispersion parameter for each cluster from choice data. 
\end{itemize}  

The solutions to the problems above can be implemented relatively easily. First, we are able to obtain $ \{\Pr(\pi_k | S)\} $ in \textit{simple} and \textit{closed-form} expressions for all $ k\geq 1 $; see \cref{thm:choice probability,thm:multiple choice probability}. This is in contrast to, say, its counterpart for the Mallows model (\cite{Antoine2021}), and we view it as our main theoretical achievement. Second, the sampling can be done in $ O(nk) $ time directly; see \cref{lma:circle}. Third, the estimation problem can be reduced to a well-studied ranking-aggregation-type formulation; see \cref{eqn:choiceAgg} and \cref{thm:ranked_choice_estimation}. Many off-the-shelf tools are available. For example, it admits a polynomial-time approximation scheme (PTAS) and can be practically solved via a linear integer programming formulation. The estimation is guaranteed to recover the model parameters asymptotically under mild conditions on the coverage of display sets; see \cref{thm:oam_consistent}. This stands in contrast to Mallows Smoothing by \cite{Antoine2021}, which cannot recover the Mallows central ranking even under sufficient coverage; see  \cref{thm:ms_unstable}. Finally, the learning problem in a mixture setting can be solved using the standard Expectation-Maximization (EM) algorithm.

We demonstrate the practical effectiveness of our methodology on two data sets on customer preference over different types of sushi. When $ k = 1 $, we compare it with two representative ranking-based choice models: one based on the Mallows and the other on Plackett-Luce (which leads to the Multinomial Logit choice model). Our tools display superb generalization power, especially when there is a limited variety of display sets in the choice data. 
When $ k > 1 $, we demonstrate the robustness of the methodology. (It is difficult to find direct comparisons for prediction power.) Specifically, as long as the underlying population keeps the same, different top-$ k $ lists aggregate into the same central rankings. 
We also show that our methodology can handle a relatively large number of items ($ n = 100 $) by achieving high precision solutions ($ < 2\% $ optimality gap) in a reasonable time (< 5 minute solving time).

\textbf{Related literature.} Our paper is most closely related to the (extensive) literature on Mallows-type ranking models and their applications to preference and choice modeling (e.g., \cite{fligner1986distance,lu2014effective,chierichetti2018mallows,vitelli2018probabilistic,Jagabathula2018,tang2019mallows, liu2019model,Antoine2021, collas2021concentric} to name a few.) Besides the new distance function and tractability results explained above, the most notable difference of this paper is the \textit{feedback structure}. In our paper, every participant chooses from an \textit{arbitrary} display set (as opposed to pairwise or the full display set only), and their feedback is in the form of a top-$k$ ranked list (as opposed to single choices, e.g.,  \cite{benson2018discrete,pfannschmidt2022learning}, or full rankings). To the best of our knowledge, the combination of both generalities makes the setting the first of its kind.\footnote{This setting is also practically relevant. For example, the platform may have capacity constraints on displaying how many items. The platform may also have an incentive to judiciously select the display sets to make feedback collection more efficient; see \cite{feng2021robust}.}  It is also worth mentioning that the analytical and computational simplicity of our model makes it a convenient building block for subsequent optimization problems (e.g., new product introduction). It is an advantage that many common methods (e.g., Monte-Carlo algorithms) do not have.

Also related is the literature on \textit{learning to rank} (e.g., \cite{cao2007learning,liu2009learning}). We wish to stress that we do not merely fit a ranking from the data but also the uncertainty quantification of the estimate (and the whole choice model $ \{\Pr(\pi_k | S)\} $). Finally, our paper is also closely related to that of \cite{feng2021robust}. They study a choice model, called the ``ordinal attraction model" (OAM), that emerges from an active learning problem of consumer preferences. We rationalize the (surprisingly simple) OAM by showing that it is equivalent to the aggregated choice probabilities $ \{\Pr(\pi_k | S)\} $ from the RMJ-based ranking model when $ k =1 $. We view it as a nontrivial observation in its own right.

\section{A Ranking Model Based on Reverse Major Index}\label{sec:RMI model}
\textbf{Preliminary.} We consider a universe of $ n $ candidates (hereafter referred to as ``items"), represented by $[n]=\{1, 2, \ldots, n\}$. We assume that every participant has a strict preference over these items, represented by a ranking (permutation). We use a bijection $\pi: [n] \rightarrow [n]$ to represent a ranking where $\pi(i)$ is the $i$th most preferred item.  We sometimes find the notation $\sigma := \pi^{-1}$ helpful, where $\sigma(x)$ is the position of item $x$ in the ranking $ \sigma $. We also use $ x \succ_{\pi} y $ if item $ x $ is preferred to item $ y $ under $ \pi $, i.e., $ \pi^{-1} (x) < \pi^{-1} (y) $. For example, $\pi = (3,1,2)$ means $ 3 \succ_{\pi} 1 \succ_{\pi} 2 $ and it corresponds to a ``$\sigma$" notation of $\sigma = (2,3,1)$. Finally, we use $e$ to represent the identity ranking, $\Sigma$ for the set of all rankings over $n$ items, $\pi_k (k \geq 1)$ for a top-$k$ list, and $\Sigma_k$ for the set of all top-$k$ rankings. 


\vspace{0.3 cm}
\textbf{Mallows-type ranking models.} Given the distance function $d(\cdot, \cdot)$, the probability mass function of the ranking for a Mallows-type model can be written as
\begin{equation*}
\lambda(\pi) = \tfrac{q^{d(\pi^*,\pi)}}{\sum_{\tpi} q^{d(\pi^*,\tpi)}},
\end{equation*}
where $\pi^*$is the central ranking, $q$ is the dispersion parameter, and $\sum_{\tpi} q^{d(\pi^*,\tpi)}$ is the normalization constant. Intuitively, the Mallows-type model defines a population of participants whose preferences are "similar" as they are centered around a common ranking, where the probability for deviations thereof decreases exponentially.

Different distance functions correspond to different models. A common requirement for a valid distance function is that it is invariant to ``relabeling." Formally, that means $d(\cdot, \cdot)$ is left-invariant under the ranking composition, i.e., $d\left(\pi_{1}, \pi_{2}\right)=d\left(\pi \pi_{1} , \pi \pi_{2} \right)$ for every $\pi, \pi_{1}, \pi_{2}$.\footnote{Equivalently, one could use the ``$\sigma$" notation and write the distance function as $\tilde{d}(\sigma_1, \sigma_2):= d(\pi_2, \pi_1)$ and $\tilde{d}$ will be right-invariant under ranking composition.\label{footenote:relabeling}} This invariance property enables to make the following conventions without loss of generality: First, we assume that the items have been properly relabeled so that the central ranking $\pi^* = e$ (unless otherwise specified). Second, we may use $d(\pi)$ as shorthand notation for $d(e, \pi)$, which fully represents a distance function with the knowledge that $ d(\pi_1, \pi_2) = d(e, \pi_1^{-1}\pi_2) $. 

\vspace{0.3 cm}
\textbf{The Mallows (Kendall's Tau distance based) ranking model.}  Commonly studied distance functions include the Spearman's rank correlation, Spearman's Footrule, and the Kendall's Tau distance (\cite{fligner1986distance}). Among those three, Kendall's Tau corresponds to the Mallows model and is defined as 
\begin{align}
d_{K}(\pi)  = \ssum_{i=1}^{n-1} \ssum_{j=i+1}^{n}  \mathbb{I}\{\pi(i)>\pi(j)\}. \label{eq:kT distance}
\end{align}

It measures a ranking's total number of pairwise disagreements (with the identity ranking $e$). For example, consider the ranking $\pi = (4,2,1,3)$. There are four pairwise disagreements: $\{(4,2), (4,1), (4,3), (2,1)\}$. Therefore, $d_{K}((4,2,1,3)) = 4$. 

As an appealing property, Kendall's tau distance leads to a tractable expression for $\lambda$. Other common distances do not have this since the normalization constant involves summing over $ n! $ rankings. The constant under the Mallows model can be expressed as 
(see \cite{marden1996analyzing}):
\begin{align}\label{eq:normalizing factor}
\ssum_{\pi} q^{d_{K}(\pi)} = \psi(n, q):= \sprod_{i=1}^{n} \tfrac{1-q^{i}}{1-q} = \sprod_{i=1}^{n} \big( 1+q+\ldots+q^{(i-1)} \big).
\end{align}

\vspace{0.3 cm}

\textbf{The RMJ-based ranking model.} In this paper, we identify a new distance function, which we call \textit{reverse major index (RMJ)}. It is defined as 
\begin{equation}
d_{R}(\pi) := \textit{Reverse Major Index} (\pi) =\ssum_{i=1}^{n-1} \, \mathbb{I}\{\pi(i)>\pi(i+1)\} \cdot(n-i) \label{eq:RMJ distance}
\end{equation}

Compared to Kendall's Tau in \eqref{eq:kT distance}, RMJ focuses on \textit{adjacent} disagreements and puts more weight on top-position disagreements. For example, consider the ranking $\pi = (4,2,1,3)$ again. The only adjacent disagreements are $\{(4,2),(2,1)\}$. Therefore, after including the weights in \eqref{eq:RMJ distance}, we have $d_{R}((4,2,1,3)) = 3 +2 = 5$. The name of RMJ is inspired by the \textit{major index} from the combinatorics literature (\cite{thanatipanonda2004inversions}), which is defined as
$
d_M(\pi)=\ssum_{i=1}^{n-1} \, \mathbb{I}\{\pi(i)>\pi(i+1)\} \cdot i.
$

\vspace{0.3 cm}

\textbf{Discussion: Kendall's Tau vs. RMJ.} Both metrics are conceptually similar: they measure a ranking's deviation from the identity. They coincide in many intuitive cases. For example, in the most extreme ones we have $d_{K} (e) =  d_{R} (e) = 0$ and $d_{K} ((n, \ldots, 1)) =  d_{R} ((n, \ldots, 1))= n(n-1)/2$. Therefore, both can be used as ``kernels" to smooth out the distribution over rankings. However, they emphasize the deviation in a subtly different way. Therefore, we can also find subtle rankings where their values are different. For example, $ d_K((4,2,1,3)) = 4 $ but $ d_R((4,2,1,3)) = 5 $. 

We believe that both Kendall's Tau and RMJ are reasonable measures of ranking discrepancy, and we find it difficult to tell which kernel is necessarily ``better" from a theoretical/axiomatic approach. For example, \cite{fligner1986distance} specifies a set of properties that a reasonable distance function should satisfy:
\begin{enumerate}
	\item The distance $d(\pi, \tilde{\pi}) \geq 0$, and the equality holds if and only if $\pi$ and $\tilde{\pi}$ are the same ranking;
	\item The distance function $d(\cdot, \cdot) $ is invariant to relabeling. (See earlier discussion in this section, especially \cref{footenote:relabeling}.)
\end{enumerate}
In this sense, both Kendall's Tau and RMJ satisfy the basic axioms for ranking distances. Meanwhile, \cite{kemeny1959mathematics} specifies a larger set of axioms so that Kendall's Tau is the unique distance function satisfying all axioms.
(For example, it can be verified that Kendall's Tau is a symmetric measure while RMJ is not.) 
However, it could also be argued that when it comes to human beings' preferences, top-position deviations matters more than bottom-position ones. If one makes that into a axiom, it will be satisfied by RMJ but not Kendall's Tau. In this regard, there is not a measure that satisfies ``all possible" axioms. %

Despite the aforementioned difficulties, we will show later that the RMJ produces a more tangible ranking model for ranked choices. For example, the RMJ-based ranking model leads to simple and estimatable (ranked) choice probabilities without compromising the desirable properties of Mallows, such as rationalizability and flexibility for a mixture setting. Moreover, regarding its ability to describe preference distributions that may occur in practice, we will demonstrate its descriptive and predictive power in a case study with real preference data. Therefore, we believe that RMJ produces a promising tool that is more tailed to the application of learning population preferences from (ranked) choice data.

\section{Analysis of the (Ranked) Choice Model $ \{\Pr(\pi_k| S) \} $}\label{sec:choice models}

In this section, we will characterize the (ranked) choice model $ \{\Pr(\pi_k| S) \} $ aggregated from the RMJ-based ranking model, formally defined as 
\begin{equation*}
\Pr(\pi_k| S)  = \ssum_{\tpi}  \lambda(\tpi) \II{\pi_k, \tpi, S },
\end{equation*}
where $ \II{\pi_k, \tpi, S } $ means the top-$ k $ list $ \pi_k $ is compatible with the ranking $ \tpi $ in the set $ S $. That is, $ \pi_k(i) \in S $ and $ \pi_k(i) \succ_{\tpi} x $ for all position $ i \in \{1, \ldots, k\} $ and item $ x \in S \setminus \{\pi_k(1), \ldots, \pi_k(i)\} $.  
Note that the summation is over rankings with nontrivial conditions. Therefore it is unclear \textit{in priori} whether \textit{any} distance-based ranking model can aggregate into a tractable $ \Pr(\pi_k| S) $ (even for $ k=1 $). We will also discuss how to estimate the ranked choice model parameters from data. In the sequel, we will write $ d(\cdot) = d_R(\cdot)$ since RMJ is the distance function of interest.

\subsection{The $ k = 1 $ case} \label{sec:top1}
When $ k=1 $, a top-$ k $ list model reduces to a choice model, which connects to a richer literature (e.g., \cite{Antoine2021}). Therefore, it is worth a separate discussion, which also helps build intuition for the $ k > 1 $ case.

\textbf{Choice probabilities.} Our main result for $ k=1 $ is summarized below, which characterizes the choice probabilities under the RMJ-based ranking model.
\begin{theorem}
	\label{thm:choice probability}
	Let a display set $S = \{x_1, x_2, \ldots, x_M\}$ be such that $x_1 < x_2 < \cdots < x_M$. Under the RMJ-based ranking model
	\begin{equation}\label{eqn:oamProb}
	Pr(\{x_i|S\})=\frac{q^{i-1}}{1+q+\ldots+q^{M-1}}.
	\end{equation}
\end{theorem}

In other words, for every display set, all the items within the display set are re-ranked so that their choice probabilities decay exponentially fast according to their \textit{relative ranking} within the display set. Noticeably, the choice probabilities in \cref{eqn:oamProb} are (much) simpler than that induced by the Mallows model, which even needs a Fast Fourier Transform to evaluate in $ O(n^2 \log n) $ time (\cite{Antoine2021}). Also, \cref{eqn:oamProb} rediscovers the ``Ordinal Attraction Model" (OAM) in \cite{feng2021robust}, which could also be viewed as a ``multiwise" generalization of pairwise noisy comparison models (e.g., \cite{braverman2008noisy,braverman2009sorting}). OAM gets its name because the ``attractiveness" (i.e., choice probability) of an item within a display set $S$ only depends on its \textit{relative} position in $S$ and therefore is only a function of the ``ordinal" information. While OAM emerges from an active learning problem of consumer preferences, we ``rationalize" it by showing that it can be aggregated from the RMJ-based ranking model, which we believe is a nontrivial observation in its own right. In the sequel, we will follow its convention and refer to the choice model defined in \cref{thm:choice probability} as OAM. 

\vspace{0.3 cm}
\textbf{Proof outline and key intermediate results for \cref{thm:choice probability}.} Let us introduce a few notations for top-$ k $ (sub-) rankings. Given a top-$ k $ ranking $ \pi_k \in \Sigma_k $, we could define RMJ for $ \pi_k $ by truncating the index at position $ k $, i.e., $d(\pi_k)=\sum_{i=1}^{k-1} \mathbb{I}\{\pi_k(i)>\pi_k(i+1)\} \cdot(n-i)$. In addition, let $ R(\pi_k) := \{\pi_k(i): i = 1, \ldots, k\} $ be the set of top-$ k $ items and $ R^c(\pi_k) := [n] \setminus R(\pi_k) $ its complement. Finally, let $ L\left(\pi_{k}\right):= |\left\{x \in R^c(\pi_k): x<\pi_{k}(k)\right\}| $ be the number of items that are (i) not included in $ \pi_k $ and (ii) having smaller indices than (i.e., preferred under $ \pi^\ast = e $) item $ \pi_k(k) $. For example, suppose $ n = 7 $ and $\pi_4 = (7,4,6,2)$. Then $ d(\pi_4) = 6+4 = 10, R(\pi_4) = \{2,4,6,7\}, R^c(\pi_4) = \{1,3,5\}, \pi_4(4) = 2, \text{and } L\left(\pi_{4}\right) = |\{1\}| = 1.$

Given two subrankings $ \pi_k \in \Sigma_k $ and  $ \pi_{k'} \in \Sigma_{k'} $ with $ k \leq k' $, we write $ \pi_k \subseteq \pi_{k'} $ if they are compatible, i.e., $ \pi_k(i) = \pi_{k'} (i) $ for all $ i = 1, \ldots, k $. Our first result concerns extending the domain of $ \lambda $ to top-$ k $ rankings, formally defined as $ \lambda(\pi_k) := \sum_{\tpi} \lambda(\tpi) \I{\pi_k \subseteq \tpi} $. 

\begin{lemma}[Probability distribution of top-$k$ rankings] \label{lma:topk_pmf}
	$\lambda \left(\pi_{k}\right)=q^{d\left(\pi_{k}\right)+L\left(\pi_{k}\right)} \cdot \frac{\psi (n-k,q)}{\psi (n,q)}$.
\end{lemma}

The significance of the result above is that we could think of a ranking $ \pi $ as a stochastic process on a ``tree" with depth $ n $ and $ n! $ leaves. While $ \{\lambda(\pi): \pi \in \Sigma\} $ describe the probability distribution over the leaves, for every $ k $, $ \{\lambda(\pi_k): \pi_k \in \Sigma_k\} $ describe the probability distribution over the nodes at level $ k $. 

As a consequence of \cref{lma:topk_pmf}, we can write out how to randomly sample a ranking $ \pi $ under the RMJ-based ranking model from the top to the bottom position. Given a top-$ k $ ranking $ \pi_k $ and an item $ z $, we write $\pi_k \oplus z$ as the concatenation of $\pi_{k}$ and item $ z $. For example, suppose $\pi_3 = (5,2,4)$, then $\pi_3 \oplus 3 = (5,2,4,3)$.


\begin{lemma}[Random ranking generation] \label{lma:circle}
	Given a top-k ranking $\pi_k $ such that $ \pi_k (k) = z $, the conditional probability for the $ (k+1) $-positioned item is
	$$
	Pr(\pi_{k+1} = \pi_k \oplus y | \pi_k) := \tfrac{\lambda(\pi_k \oplus y)} {\lambda(\pi_k)}= \tfrac{q^{h(y|z) - 1}}{1+q+\dots+q^{n-k-1}}, 
	$$    
	where $h(y|z) =                                                                            
	\begin{cases}
	\sum_{x \in R^c(\pi_k)} \mathbb{I} \{z < x \leq y\} &   \text{  if } y>z \\
	n - k - \sum_{x \in R^c(\pi_k)} \mathbb{I} \{y<x<z\}& \text{  if } y<z
	\end{cases}.
	$
\end{lemma}

Finally, \cref{lma:topk_pmf,lma:circle} lead to the following result based on an induction argument.

\begin{lemma} \label{lma:oam_closed_form}
	Let a display set $S = \{x_1, x_2, \ldots, x_M\}$ be such that $x_1 < x_2 < \cdots < x_M$ and a top-$ k $ ranking $ \pi_k $ be such that $ \pi_k = \pi_{k-1} \oplus z $ and $ R( \pi_{k-1}) \cap S = \emptyset$. Then conditional on a participant's top-$ k $ preference list is $ \pi_k $, the probability that (s)he will choose item $ x_i $ out of display set $ S $ is
	\begin{equation*}
	Pr(\{x_i|S\}|\pi_k) := \frac{\ssum_{\tpi} \lambda(\tpi) \cdot \II{x_i, \tpi, S } \cdot \I{\pi_k \subseteq \tpi}}{\lambda(\pi_k)} =  
	\begin{cases}
	\tfrac{q^{M-p(z|S)+i-1}}{1+q+\ldots+q^{M-1}} & \text{ if }  z > x_i \text{ and } z \not\in S , \\
	\noalign{\vskip5pt}
	\tfrac{q^{i-p(z|S)-1}}{1+q+\ldots+q^{M-1}} & \text{  if }  z < x_i  \text{ and } z \not\in S , \\
	\noalign{\vskip3pt}
	1 & \text{  if } z = x_i, \\
	0 & \text{  if } z \in S \setminus \{x_i\}. 
	\end{cases}
	\end{equation*}
	where $p(z|S) := \sum_{x \in S} \mathbb{I} \{x < z\}.$
\end{lemma}

Note that \cref{lma:oam_closed_form} could be viewed as a generalization of \cref{thm:choice probability} by setting $ \pi_k = \emptyset $ (which corresponds to $ z = 0 $).

\vspace{0.3 cm}

\textbf{Parameter learning from choice data.} In the parameter learning problem, we are endowed with choice data $ H_T = (S_1, x_1, \ldots, S_T, x_T) $, where $ S_t $ is the display set shown to the $ t $th participant and $ x_t $ is his/her choice. 
Following \cite{feng2021robust}, the maximum likelihood estimator (MLE) for the central ranking $ \pi^\ast $ can be obtained from the following \textit{choice aggregation} problem
\begin{align*}
\hat{\pi} \in \text{argmin}_{\pi} \ssum_{t=1}^{T} \ssum_{x \in S_t} \I{x \succ_\pi x_t}.
\end{align*}
It has a further linear integer programming formulation. Let $w_{i j} :=\sum_{t=1}^{T} \mathbb{I}\left\{\{i, j\} \subseteq S_{t}\right.$ and $\left.x_{t}=i\right\}$ be the number of times that item $i$ and item $j$ are displayed together and item $i$ is chosen. Intuitively, a positive $w_{i j}-w_{j i}$ is an indication that item $ i $ should be preferred to item $ j $. Invoking Proposition 3 in \cite{feng2021robust}, the integer programming could be written as follows:

\begin{equation}\label{eqn:choiceAgg}
\begin{aligned}
\hat{x} \in  \text{argmin}_{x} \quad &\ssum_{(i,j): i \neq j} w_{ij} x_{ji}\\
\textrm{s.t.} \quad & \begin{array}{ll}
x_{i j}+x_{j i}=1 & \forall 1 \leq i, j \leq n, i \neq j \\
x_{i j}+x_{j r}+x_{r i} \leq 2 & \forall 1 \leq i, j, r \leq n, i \neq j \neq r \\
x_{i j} \in\{0,1\} & \forall 1 \leq i, j \leq n
\end{array}
\end{aligned}
\end{equation}
\vspace{0.1cm}

In the formulation above, the solution $\hat{x}$ is such that $\hat{x}_{i j}=1$ if $i \succ_{\hat{\pi}} j$ under the MLE $ \hat{\pi} $. Computationally, this integer programming is an instance of the well-studied \textit{feedback arc set problem on tournaments}. Therefore, it admits a polynomial-time approximation scheme (PTAS). From a more practical side, the central ranking $ \pi^\ast $ can be effectively obtained using off-the-shelf integer programming solvers and with many speeding-up heuristics; see \cite{feng2021robust} for more discussion. The MLE for dispersion parameter $ q $ can be subsequently obtained from an one-dimensional (convex) optimization problem $ \hat{\alpha} \in \text{argmin}_{\alpha} \{ \alpha  \ssum_{t=1}^{T} \ssum_{x \in S_t} \I{x \succ_{\hat{\pi}} x_t} + \sum_{t=1}^{T} \log {\sum\nolimits_{j=0}^{|S_t|-1} e^{-j\alpha} } \}$ so that  $ \hat{\alpha} = - \ln \hat{q}  $. It is also worth noting that the MLE framework can be easily extended to learning in a mixture setting using the standard EM algorithm, which we refer to the supplementary materials for more details.

Let us conclude this section by providing some theoretical understanding of whether we could recover the ground truth values of $ \pi^\ast $ and $ q $ at least asymptotically. Intuitively, this should depend on the ``coverage" of display sets: if only a pair $ \{x,y\} $ is repeatedly, there is no hope of recovering the full ranking $ \pi^\ast $. It turns out that the parameters can be recovered as long as every pair is ``covered" by some display set.

\begin{theorem}\label{thm:oam_consistent}
	Consider a sequence of display sets $ \{S_t\} $. Suppose every pair of items $ \{i,j\} \subseteq [n]$ is displayed infinitely often. That is, $ \sum_{t=1}^{T} \mathbf{I} \big\{\{i,j\in S_t\} \big\} \to \infty $ as $ T \to \infty $. Then the MLE $ (\hat{\pi}, \hat{q}) = (\hat{\pi}(H_T), \hat{q}(H_T)) $ is an consistent estimator. That is, $ (\hat{\pi}, \hat{q}) \to ({\pi}^\ast, q^\ast) $ almost surely as $ T \to \infty $. Conversely, if there exists a pair of items $ \{i, j\} $ that is only displayed finitely often, then there exists a tie-breaking rule for MLE so that $ \hat{\pi} \not\to {\pi}^\ast $ with positive probability.
\end{theorem}
We would like to mention that the coverage condition is fairly mild. For example, as long as the full display set is displayed sufficiently many times, the RMJ-implied central ranking can eventually be recovered. This stands in contrast to its counterpart in the Mallows model. Since the Mallows model does not produce simple choice probabilities, the MLE from choice data is rather difficult to obtain. Perhaps the best method to date is by \cite{Antoine2021}, who uses a ``Mallows smoothing" heuristic. The following result reveals that the heuristic can be unstable or fail to recover the underlying Mallows parameters even under sufficient coverage of display sets.

\begin{theorem}\label{thm:ms_unstable}
	Even if all display sets with sizes larger than two are displayed infinite times, the estimator from the Mallows Smoothing heuristic is not consistent.  
\end{theorem}

The intuition behind the result above is that Mallows Smoothing needs a ``choice to ranking" step. That is, it needs to first find a distribution over rankings, denoted by $ \hat{\lambda} \in \Delta(\Sigma) $, that aggregates into choice probabilities to match the empirical choice probabilities $ \{ \hat{\Pr}(x|S)\} $. Formally, it corresponds to solving the system of linear equations
\begin{equation}\label{eq:system of equations}
\ssum_{\tpi}  \hat{\lambda}(\tpi) \II{x, \tpi, S} = \hat{\Pr}(x|S)
\end{equation}

for all display set $ S $ in the data and $ x \in S $. Since there are vastly more variables than equations, the solutions in general form a polytope rather than a singleton. More importantly, we find that there can be solutions $ \hat{\lambda}_1 $ and $ \hat{\lambda}_2 $ that aggregate to different Mallows models in the ``ranking aggregation" step of the heuristic. Therefore, in general, this heuristic can produce non-unique results and thus be inconsistent. 

\subsection{The general ($ k \geq 1 $) case} \label{sec:topk}

In the general ($ k \geq 1 $) case, the ranked choice probabilities and the corresponding estimation remain parsimonious. Roughly speaking, the ranked choice model behaves like a generalized OAM where $ \Pr(\pi_k | S) $ depends on the \textit{relative rankings} of the items in $ \pi_k $ among the set $ S $. The tractability comes from the sole dependence on the \textit{relative} ranking. That is, to obtain $ \Pr(\pi_k| S) $, one could first treat $ S $ as the ``full display set" for a subuniverse of items, then re-rank all the items within the display set $ S $, and finally just apply \cref{lma:topk_pmf}. 

\vspace{0.3 cm}

\textbf{Choice probabilities.} Given a display set $ S $, let $d_S(\cdot)$, $L_S(\cdot)$ be the originally defined $ d(\cdot) $ and $ L(\cdot) $ functions, but treating the display set $ S $ as the  universe. Formally, $ d_S(\pi_k) := \sum_{i=1}^{k-1} \mathbb{I}\{\pi_k(i)>\pi_k(i+1)\} \cdot(|S|-i) $ and $ L_S\left(\pi_{k}\right):= |\left\{x \in R^c(\pi_k) \cap S: x<\pi_{k}(k)\right\}| $. For example, suppose again that $ n = 7 $ and $\pi_4 = (7,4,6,2)$, and let $S = \{2,3,4,5,6,7\}$. Recall that $ d(\pi_4) = 6+4 = 10 \text{ and } L\left(\pi_{4}\right) = |\{1\}| = 1.$ In comparison, $ d_S(\pi_4) = 5+3 = 8 \text{ and } L_S\left(\pi_{4}\right) = |\emptyset| = 0.$

\begin{theorem}\label{thm:multiple choice probability}
	Given a display set $ S $ and a top-$ k $ ranking $ \pi_k $ such that $ R(\pi_k) \subseteq S $, we have
	\begin{equation*}
	Pr(\pi_{k}|S) = q^{d_S(\pi_{k})+L_S(\pi_{k})} \cdot \frac{\psi (|S|-k,q)}{\psi (|S|,q)}.
	\end{equation*}
\end{theorem}

Note that \cref{thm:multiple choice probability} generalizes both \cref{thm:choice probability} and \cref{lma:topk_pmf} (but in different ways). It reduces to \cref{thm:choice probability} by setting $ k =1 $: Given $ \pi_1 = (z) $ where $ z \in S $, we have $ d_S(\pi_{1}) = 0$, $ L_S(\pi_{1}) = |\{x \in S \setminus \{z\} : x < z\}| $ the relative ranking of $ z $ in display set $ S $, and $ \tfrac{\psi (|S|-k,q)}{\psi (|S|,q)} = 1 + q +\cdots + q^{|S|-1} $ the appropriate normalizing constant where $ k = 1 $. In addition, \cref{thm:multiple choice probability} reduces to \cref{lma:topk_pmf} by setting $ S = [n] $, as can be easily verified.

\vspace{0.3 cm}

\textbf{Parameter learning from ranked choice data.} In the parameter learning problem, we are endowed with choice data $ H_T = (S_1, \pi_{k}^{1}, \ldots, S_T, \pi_{k}^{T}) $, where $ S_t $ is the display set shown to the $ t $th participant and $ \pi_{k}^{t} = (x_1^t, \ldots, x_k^t) $ are his/her top-$ k $ choices ranked from the most preferred to the $ k $th preferred.

It turns out that (somewhat surprisingly) in the $ k \geq 1  $ case, the central ranking $ \pi^\ast $ can be estimated from the same integer programming \eqref{eqn:choiceAgg} but just with a generalized re-definition of the weight parameters $ \{w_{ij}\} $.
\begin{theorem}\label{thm:ranked_choice_estimation}
	The MLE for the central ranking $ \pi^\ast $, $ \hat{\pi} $, can be obtained from the integer program \eqref{eqn:choiceAgg} with $\hat{x}_{i j}=1$ iff $i \succ_{\hat{\pi}} j$ and a generalized definition of $ \{w_{ij}\} $ below
	\begin{equation*}
	\begin{split}
	w_{ij} = \sum_{t=1}^{T}  \bigg[ \mathbb{I}\{ x_k^t = i\} \cdot \mathbb{I} \big\{ \{i, j \} \subseteq S_t \backslash \{x_1^t,\ldots,x_{k-1}^t\} \big\} +  \sum_{h=1}^{k-1} (|S_t|-h) \cdot \mathbb{I}\{ x_h^t = i, x_{h+1}^t = j\}     \bigg]
	\end{split}.
	\end{equation*}
\end{theorem}

Intuitively, a positive $w_{i j}-w_{j i}$ is still an indication that item $ i $ should be preferred to item $ j $, but now taking into consideration that every participant's response is actually a \textit{ranked list} rather than a single choice. Practically speaking, the $ \{w_{ij}\} $ can be maintained in a relatively simple manner. Suppose a display set $ S $ and a ranked choice $ \pi_k $ are given. Then $ w_{\pi_k(\ell), \pi_k(\ell+1)} $ will be added by $ |S|-\ell $ for all $ \ell = 1, \ldots, k-1 $. (This captures the ranking information for items included in $ \pi_k $.) In addition, $ w_{\pi_k(k), j} $ will be added by $ 1 $ for all $\ell \in \{R^c(\pi_k) \cap S \}$. (This captures the ranking information between $ \pi_k(k) $ and all items that are not included in $ \pi_k $.) For example, suppose $n=6$ and $k=3$. The ranked choice data is such that $ S_1 = \{1,2,3,4,5\} \text{ and } \pi_k^1 = (3,1,2)$. Then we have $w_{3,1}=4, w_{1,2}=3, w_{2,4}=w_{2,5}=1 \text{ and all the other } w_{ij} = 0.$ 

\vspace{0.3 cm}
\textbf{Discussion: model limitation.} A few more words on the ranked choice model $ \{Pr(\pi_{k}|S)\} $. The specific form gives it great simplicity, but as with any other parametric model, it may be vulnerable to model missspecification issues in practice. In particular, since the model is calculated from a unimodal ranking distribution, this model (in its ``vanilla" form) is best suited when there is an approximate ``consensus" ranking in the population. In our numerical studies, we will demonstrate how we can effectively mitigate those issues by considering learning the model parameters in a mixture setting to better capture the preference heterogeneity in the population.

\section{Numerical Experiments} \label{numerical}
We investigate the performance of our methodology on two anonymous survey data sets regarding sushi preferences (\cite{Kamishima2015}): The first one consists of 5,000 full preference rankings over 10 kinds of sushi, while the second one consists of 5,000 top-$10$ rankings over 100 kinds.\footnote{In the supplementary material, we also conducted a numerical experiment on an e-commerce dataset, and the result is consistent with experiment 1 on sushi data.} Throughout the numerical experiments, we use a workstation with dual Intel Xeon Gold 6244 CPU (3.6 GHz and 32 cores in total), no GPUs, and 754 GB of memory.

\textbf{Experiment 1: Top-$1$ choice.} When $k=1$, a top-$ k $ list corresponds to a choice model, which connects to a richer literature (e.g., \cite{luce1959individual,Antoine2021,feng2021robust}). We compare our prediction power with two representative ranking models: Mallows and Plackett-Luce. (The previous is distance-based. The latter is not but leads to the famous multinomial logit model (MNL) choice model.) We use the Mallows Smoothing (MS) heuristic to estimate a Mallows model and MLE to estimate a Plackett-Luce. For all three rankings models (Mallows, PL, and ours), we also use an EM algorithm to perform parameter learning in a mixture setting.

Since the estimation and generalization performances can depend on the display sets, let us define an instance of the comparison by $ \{\calS_{train}, \calS_{test}, \calC \} $. Here $ \calS_{train} $ (resp. $ \calS_{test} $) is collection of display sets in the training (resp. test) data, and $ \calC $ is the number of clusters. In every instance, we split the data into 4000 participants in training and 1000 in testing. We generate empirical training (resp. testing) choice data by first enumerating all display sets in $ \calS_{train} $ (resp. $ \calS_{test} $) and then record every participant's favorite type of sushi for every display set in the training (resp. testing) data. Finally, we use the log-likelihood as performance metric. 
We will use term \textit{explanation power} (resp. \textit{prediction power}) to refer the performance metric evaluated on the training (resp. testing) data.

We use two configurations for $ \calS_{train} $ and $ \calS_{test} $, respectively. For $ \calS_{train} $, we take it to be either the full set or a collection of three randomly generated sets, which are realized to be $ \{\{1,3,4,7,8,10\}, \{2,4,5,6,8,9\}, [10]\} $. For $ \calS_{test} $, we take it to be either the collection of all pairwise sets or all display sets with sizes at least two. Note that $ \calS_{train} $ has low variability and many elements in $ \calS_{test} $ do not appear in $ \calS_{train} $, making the experiment emphasizing more on the \textit{generalization power}. We take $ \calC \in \{1,2,4,6,15\} $.

\begin{figure}[htbp]
	\centering
	\caption{Comparison of Explanation and Prediction Power for Top-$1$ choice}
	\includegraphics[width=1 \textwidth]{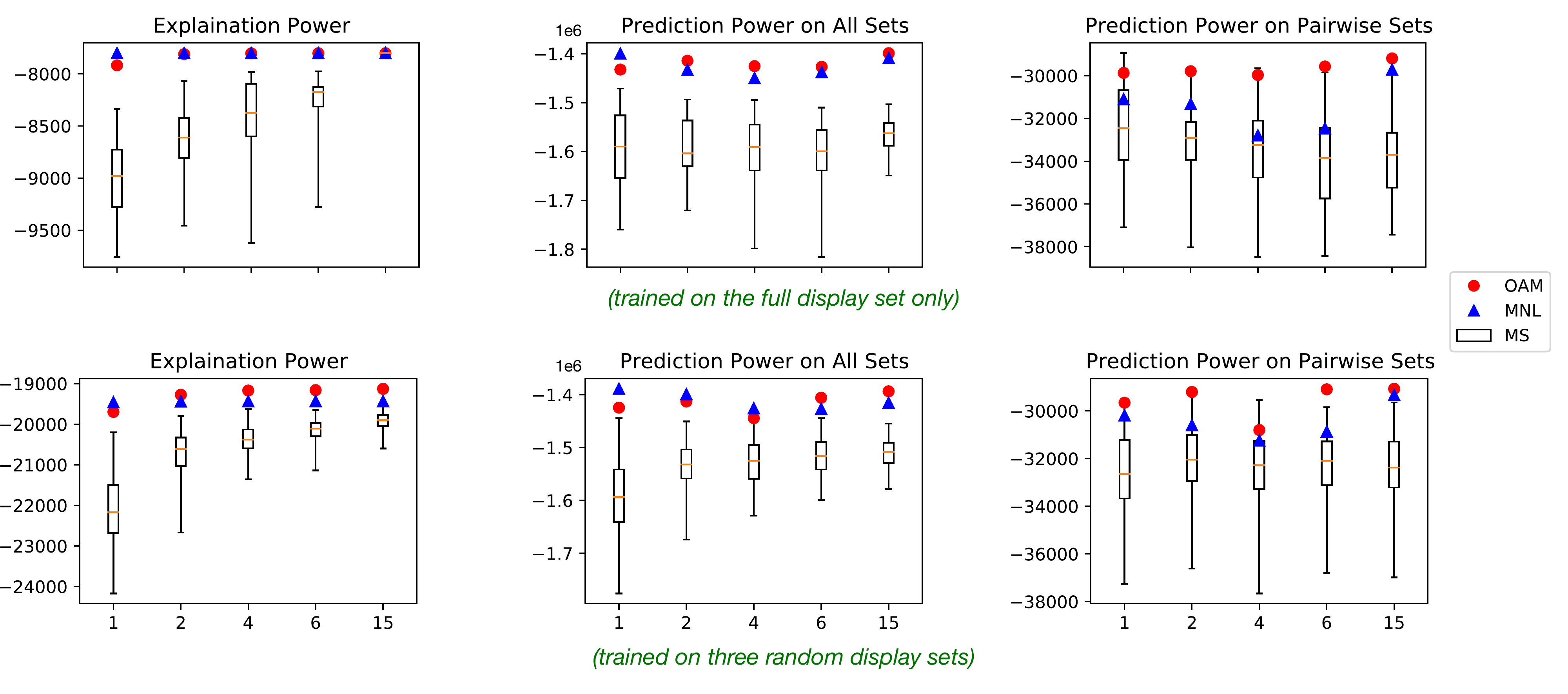}
	\ \\
	{\scriptsize \sf In each panel, the x-axis represents the number of clusters, and the y-axis represents the log-likelihood metric.}
	\label{explanation_predictionPower_n10_nlto2_l0}
\end{figure}

We summarize our results in \cref{explanation_predictionPower_n10_nlto2_l0}. We find that our method has favorable performance compared to MS and MNL across the settings. Intuitively, MS and MNL underperform in different ways. MS is vulnerable to the identifiability issue in the ``choice-to-ranking" step, for which we provide theoretical understanding in \cref{thm:ms_unstable}. Our numerical experiments confirm the theoretical insight: We randomize over the extreme points of the polytope of solutions to \eqref{eq:system of equations}, each of which leads to a valid output of the MS heuristic. Every box plot in \cref{explanation_predictionPower_n10_nlto2_l0} contains a summary of the performances of those outputs. It is clear that the performances (both in training and testing) span wide ranges. In the meantime, MNL is vulnerable to both the model risk from its specific parametric form (reflected in the limited increase of explanation power when considering more clusters) and the sample risk of overfitting (reflected in the difference in performance in training vs. testing, especially when trained on the full display set only); see more details in the supplementary materials.

\vspace{0.2 cm}

\textbf{Experiment 2: Top-$k$ choice.} In this experiment, we perform a robustness check based on the criterion that a good model should return similar central rankings under different feedback structures, i.e., different $ k $. Specifically, we conduct estimation on top-$1$, top-$2$ and top-$3$ ranked choices constructed from the first 10-sushi data set. The collection of display sets is taken to be all display sets with sizes at least $ k $. We find that all three experiment instances produce the same estimated RMJ-implied central ranking $ \hat{\pi} = (8, 5, 6, 3, 2, 1, 4, 9, 7, 10) $. Such robustness stands in contrast with other more naive methods, such as various versions of Borda count; see more details in the supplementary materials. We believe it presents evidence that our methodology is learning sensible information from the data. 



%
%

\vspace{0.2 cm}
\textbf{Experiment 3: 100 sushi types.}
In this experiment, we wish to show how effective our method is for a relatively large number of items. In the data, each of the 5000 individuals indicates their top-$ 10 $ choices out of 100 types of sushi. We use the {\sf LP-Rand-Pivot} speed-up heuristic by \cite{feng2021robust} based on LP relaxation to train a single-cluster RMJ-based model. We bootstrap 10 times (each time drawing 10000 samples) and record the running times and optimality gaps in the \cref{tab:computation table}. We find that we can obtain $ < 2\% $ optimality gap within 5 minutes (excl. model building time).

\begin{table}[htbp]
	\centering
	\caption{Computational Time and Optimality Gap on the 100-sushi data}
	\vspace{-0.2 cm}
	\label{tab:computation table}
	\begin{tabular}{cccc}
		\hline
		& Model Building Time (min) & Model Solving Time (min) & Optimality Gap \\ \hline
		Average & 21.10                     & 4.20                     & 1.47\%         \\ \hline
		Max     & 21.19                     & 4.50                     & 1.79\%         \\ \hline
	\end{tabular}
\end{table}






\vspace{-0.1 cm}
\section{Conclusions}

We identify a novel distance-based (Mallows-type) ranking model. It aggregates into simple probability distributions for top-$k$ subrankings among an arbitrary display set $S$. In addition, it facilitates effective parameter learning through the MLE formulation. This is the first distance-based ranking model with such properties (even for $k=1$) to the best of our knowledge.  

This ranking model can be used to model population preferences and provide a rationalizable way to model their ranked choices from given display sets. We demonstrate its practical value using real preference data. For example, under a mixture setting with only a few clusters, it shows promising prediction power, especially when there is a limited variety in the display sets. 

For future steps, we believe our work can serve as the ``infrastructure" for a range of business-related decision problems, such as new product introduction, crowdsourcing, and marketing research, among others.

\begin{ack}
This work is supported by the Singapore Ministry of Education (MOE) Academic Research Fund (AcRF) Tier 1 [WBS Number: R-314-000-121-115]. The authors would like to thank Ren\'{e} Caldentey and Christopher Thomas Ryan for earlier discussions of this problem, as well as Long Zhao and Xiaobo Li for their feedback. The authors are also very grateful to the Area Chair and reviewers for their careful reading of the paper and for the many helpful and constructive comments. 
 \end{ack}

\newpage

\bibliographystyle{unsrt}
\bibliography{ref}

\newpage


\newpage
\appendix

\centerline{\large Supplemental Material for:} 

\vspace{0.2 cm}

\centerline{\textit{\Large On A Mallows-type Model For (Ranked) Choices}}

\section{Characterizing the Choice Probability (\cref{thm:choice probability})}

\subsection{Proofs of Intermediate Results (\cref{lma:topk_pmf,lma:circle,lma:oam_closed_form})}

\begin{proofref}{\cref{lma:topk_pmf}}
	Let us first assign a weight to each ranking $ \pi \in \Sigma $ and subranking $ \pi_{k} \in \Sigma_{k} $ ($k \in\{0, \ldots, n\}$) by 
	\begin{align}\label{eq:definition of w}
	w(\pi) := \frac{q^{d(\pi)}}{\psi(n,q)} \quad \text{ and } \quad w\left(\pi_{k}\right) :=\sum_{\pi \in \Sigma: \pi_{k} \subseteq \pi} w(\pi).
	\end{align}
	We note that on the one hand, $ w \propto \lambda $. On the other hand, it is unclear \textit{in priori} whether $ w(\cdot) $ is the probability mass function since one needs to verify that $ w(\cdot) $ is equipped with the right normalizing constant. We will first prove that $ w(\pi_k) $ can also be written as $w\left(\pi_{k}\right)=q^{d\left(\pi_{k}\right)+L\left(\pi_{k}\right)} \cdot \frac{\psi (n-k,q)}{\psi (n,q)}$. Then we will show that for every $ k \in [n] $, $ \sum_{\pi_k \in \Sigma_{k}} w(\pi_k) = 1 $. 
	In other words, $ w (\cdot) = \lambda(\cdot) $.

	\underline{Step 1.} Recall that for every $ \pi_k \in \Sigma_{k} $, $d(\pi_k)=\sum_{i=1}^{k-1} \mathbb{I}\{\pi_k(i)>\pi_k(i+1)\}(n-i)$ and $ L\left(\pi_{k}\right):= |\left\{x \in R^c(\pi_k): x<\pi_{k}(k)\right\}| $. We prove from \eqref{eq:definition of w} that
	$$w\left(\pi_{k}\right)=q^{d\left(\pi_{k}\right)+L\left(\pi_{k}\right)} \cdot \frac{\psi (n-k,q)}{\psi (n,q)}$$ by backward induction on $k$.
	
	\emph{Base step.} Suppose $k=n.$ Pick an arbitrary $\pi_{k} \in \Sigma_{k}=\Sigma .$ Because $\pi_{k}$ is a full ranking,
	\begin{align*}
	w(\pi_{k}) = q^{d(\pi_{k})} \cdot \frac{1}{\psi (n,q)} \stackrel{(a)}{=} q^{d(\pi_{k})+L(\pi_{k})} \cdot \frac{\psi (n-k,q)}{\psi (n,q)} .
	\end{align*}
	
	In the derivations above, part (a) is due to the facts that $\psi (n-k,q)=\psi (0,q)=1$ and $L(\pi_{k})=0$ by definition.
	
	\emph{Inductive step.} Pick an arbitrary $K \in\{0,1, \ldots, n-1\}$. Suppose our statement holds for every $k=K+1, \ldots, n$. We want to show that our statement holds for $k=K$. In other words, pick an arbitrary $\pi_{k} \in \Sigma_{k}$. We wish to show that $\Sigma_{k}, w(\pi_{k})=q^{d(\pi_{k})+L(\pi_{k})} \cdot \frac{\psi (n-k,q)}{\psi (n,q)}.$
	
	We first claim that for every $ \pi_{k+1} \in \Sigma_{k+1} $ such that $\pi_k \subseteq \pi_{k+1}$,  $\pi_{k}(k)>\pi_{k+1}(k+1)$ if and only if $L(\pi_{k+1}) \leq L(\pi_{k})-1$. To see why, note that if $\pi_{k}(k)>\pi_{k+1}(k+1)$, $L(\pi_{k}) = L(\pi_{k+1}) + 1$. Otherwise, $L(\pi_{k}) \leq L(\pi_{k+1})$, and hence $L(\pi_{k}) -1 < L(\pi_{k+1})$. As a consequence, $d\left(\pi_{k+1}\right)$ can be expressed in terms of $d\left(\pi_{k}\right)$ :
	\begin{align*}
	d\left(\pi_{k+1}\right) &= \sum_{i=1}^{k} \mathbb{I}\left\{\pi_{k}(i)>\pi_{k}(i+1)\right\} \cdot(n-i) \\
	&= \underbrace{\sum_{i=1}^{k-1} \mathbb{I}\left\{\pi_{k}(i)>\pi_{k}(i+1)\right\} \cdot(n-i)}_{d\left(\pi_{k}\right)}+\underbrace{\mathbb{I}\left\{\pi_{k}(k)>\pi_{k}(k+1)\right\}}_{\mathbb{I}\left\{L\left(\pi_{k+1}\right) \leq L\left(\pi_{k}\right)-1\right\}} \cdot(n-k) \\
	&= d\left(\pi_{k}\right)+\mathbb{I}\left\{L\left(\pi_{k+1}\right) \leq L\left(\pi_{k}\right)-1\right\} \cdot(n-k) .
	\end{align*}
	
	Moreover, given $\pi_{k}$, there is a one-to-one correspondence between $\pi_{k+1}$ and ${k+1}$ and we use $\pi_{k}\oplus L^{-1}(i)(k+1)$ to represent the (unique) $\pi_{k+1}$ such that $\pi_{k} \subseteq \pi_{k+1}$ and $L(\pi_{k+1})=i, (i \in \{0,1,\ldots,n-k-1\})$.
	
	\begin{align*}
	w\left(\pi_{k}\right)&=\sum_{\pi \in \Sigma: \pi_{k} \subseteq \pi} w(\pi)=\sum_{\pi_{k+1} \in \Sigma_{k+1}: \pi_{k} \subseteq \pi_{k+1}} w\left(\pi_{k+1}\right)\\
	&=\sum_{i=0}^{n-k-1} w\left(\pi_{k}\oplus L^{-1}(i)(k+1)\right) \\
	&=\sum_{i=0}^{n-k-1} q^{d\left(\pi_{k}\oplus L^{-1}(i)(k+1)\right)+L\left(\pi_{k}\oplus L^{-1}(i)(k+1)\right)} \cdot \frac{\psi (n-k-1,q)}{\psi (n,q)}\\
	&=\frac{\psi (n-k-1,q)}{\psi (n,q)} \cdot\left(\sum_{i=0}^{L\left(\pi_{k}\right)-1} q^{d\left(\pi_{k}\oplus L^{-1}(i)(k+1)\right)+i}+\sum_{i=L\left(\pi_{k}\right)}^{n-k-1} q^{d\left(\pi_{k}\oplus L^{-1}(i)(k+1)\right)+i}\right) \\
	&=\frac{\psi (n-k-1,q)}{\psi (n,q)} \cdot\left(\sum_{i=0}^{L\left(\pi_{k}\right)-1} q^{d\left(\pi_{k}\right)+n-k+i}+\sum_{i=L\left(\pi_{k}\right)}^{n-k-1} q^{d\left(\pi_{k}\right)+i}\right) \\
	&=\frac{\psi (n-k-1,q)}{\psi (n,q)} \cdot q^{d\left(\pi_{k}\right)} \cdot q^{L\left(\pi_{k}\right)}\underbrace{\left(\sum_{i=n-k-L\left(\pi_{k}\right)}^{n-k-1} q^{i}+\sum_{i=0}^{n-k-1-L\left(\pi_{k}\right)} q^{i}\right)}_{\frac{\psi (n-k,q)}{\psi (n-k-1,q)}} \\
	&=q^{d\left(\pi_{k}\right)+L\left(\pi_{k}\right)} \cdot \frac{\psi (n-k,q)}{\psi (n,q)}.
	\end{align*}
	
	Hence our statement holds for $k = K$, too, thus $w\left(\pi_{k}\right)=q^{d\left(\pi_{k}\right)+L\left(\pi_{k}\right)} \cdot \frac{\psi (n-k,q)}{\psi (n,q)}$ for any $k$. 
	
	\vskip 0.2 cm
	
	\underline{Step 2.} We verify that for every $ k \in \{0, 1, \ldots, n\} $, $ \sum_{\pi_k \in \Sigma_{k}} w(\pi_k) = 1 $.
	
	For all $ k, k'  \in \{0, 1, \ldots, n\}$, we invoke \eqref{eq:definition of w} and have $\sum_{\pi_k \in \Sigma_k} w(\pi_k) = \sum_{\pi \in \Sigma} w(\pi) = \sum_{\pi_{k^{\prime}} \in \Sigma_{k^{\prime}}} w(\pi_{k^{\prime}})$. Therefore, take an arbitrary $k  \in \{0, 1, \ldots, n\} \text{ and } k^{\prime}=2$, 
	\begin{align*}
	\sum_{\pi_{k} \in \Sigma_{k}} w(\pi_{k}) &= \sum_{\pi_{2} \in \Sigma_{2}} w(\pi_{2}) = \sum_{i \in \{0,\ldots,n-1\}} w(\pi_{1}\oplus i)\\
	&= \frac{\psi (n-1,q)}{\psi (n,q)} \cdot (1+q+\ldots+q^{n-1})\\
	&= 1
	\end{align*}
	Hence, the weight $w$ is the probability mass function $\lambda$. 
\end{proofref}

\vspace{0.4 cm}

\begin{proofref}{\cref{lma:circle}}
	By Lemma \ref{lma:topk_pmf}, we have
	\begin{equation*}
	\begin{aligned}
	\frac{\lambda(\pi_k \oplus y)} {\lambda(\pi_k)}
	=& \frac{q^{d\left(\pi_{k+1}\right)+L\left(\pi_{k+1}\right)} \cdot \frac{\psi (n-k-1,q)}{\psi (n,q)}}{q^{d\left(\pi_{k}\right)+L\left(\pi_{k}\right)} \cdot \frac{\psi (n-k,q)}{\psi (n,q)}}
	= \frac{q^{d(\pi_k)+\mathbb{I} \{z>y\}\cdot (n-k)+L(\pi_{k+1})}}{q^{d(\pi_k)+L(\pi_k)}}\frac{1-q}{1-q^{n-k}}\\
	=& q^{\mathbb{I} \{z>y\}\cdot (n-k)+L(\pi_{k+1})-L(\pi_k)}\frac{1-q}{1-q^{n-k}}.
	\end{aligned}
	\end{equation*}
	By the definition of $L(\cdot)$, 
	$$
	\mathbb{I} \{z>y\}\cdot (n-k)+L(\pi_{k+1})-L(\pi_k) = 
	\begin{cases}
	\sum_{x \in R^c(\pi_k)} \mathbb{I} \{z < x \leq y\} - 1 &   \text{  if } y>z \\
	n - k - \sum_{x \in R^c(\pi_k)} \mathbb{I} \{y<x<z\} - 1& \text{  if } y<z
	\end{cases}. 
	$$
	The proof is finished.
\end{proofref}

\vspace{0.4 cm}

\begin{proofref}{\cref{lma:oam_closed_form}}
    We prove this lemma by backward induction on $k$.
    
    \emph{Base step.} Since when $k>n-M+1$, $ R( \pi_{k-1}) \cap S = \emptyset$ is not satisfied for all $\pi_{k-1} \in \Sigma_{k-1}$. We suppose $k=n-M+1$. Pick an arbitrary $\pi_{k}=\pi_{k-1} \oplus z$ such that $ R( \pi_{k-1}) \cap S = \emptyset$. Item $ z $ is an item in $S$, so the choice outcome is deterministic. If $ z = x_i $, the conditional choice probability is 1, and if $ z \in S \setminus \{x_i\} $, the conditional choice probability is 0. 
    
    \emph{Inductive step.} Pick an arbitrary $K \in\{1, \ldots, n-M\}$. Suppose our statement holds for every $k=K+1, \ldots, n-M+1$. We want to show that our statement holds for $k=K$. 
	
	Pick an arbitrary $\pi_{k}=\pi_{k-1} \oplus z$ such that $ R( \pi_{k-1}) \cap S = \emptyset$. Similar to the base step, if $ z = x_i $, the conditional choice probability is 1, and if $ z \in S \setminus \{x_i\} $, the conditional choice probability is 0. When $z \notin S$, rename the items in $R^c( \pi_{k})$ as $y_1,y_2,\ldots,y_{n-k}$ so that 
	$y_1 < y_2 < \ldots < y_{n-k}$. 
	We have the following decomposition.
	\begin{equation*}
	Q:= Pr(\{x_i|S\}| \pi_{k-1} \oplus z) = 
	\sum_{j=1}^{n-k} Pr(\{x_i|S\}|\pi_{k} \oplus y_j) \cdot Pr(\pi_{k} \oplus y_j |\pi_{k-1} \oplus z) 
	\end{equation*} 
	Since $ R( \pi_{k}) \cap S = \emptyset$, we have $\{x_1, x_2, \ldots, x_M\} \subseteq \{y_1, y_2, \ldots, y_{n-k}\}$, we use $\overline{x_i}$ to denote the relative position of item $x_i$ in these $n-k$ items, hence item $x_i$ is renamed as $y_{\overline{x_i}}$ under ``y" notation. For example, if $\overline{x_i}=4$, then there are 3 items with smaller indices (i.e., preferred under $\pi^*=e$) than item $x_i$ in $\{y_i\}_{i=1}^{n-k}$. Hence, we can rewrite $\{y_i\}_{i=1}^{n-k}$ as 
	$$\{y_1, \ldots, y_{\overline{x_1}-1}, y_{\overline{x_1}}, y_{\overline{x_1}+1}, \ldots, y_{\overline{x_i}-1}, y_{\overline{x_i}}, y_{\overline{x_i}+1}, \ldots, y_{n-k}\}.$$

	Because different $\pi_{k+1}(k+1)$ could result in the same $p(\pi_{k+1}(k+1)|S)$, we classify $\pi_{k+1}(k+1)$ by its corresponding $p(\pi_{k+1}(k+1)|S)$ and by induction hypothesis, we have
	
	\begin{align*}
	 \Pr(\{x_i|S\}|\pi_{k} \oplus \pi_{k+1}(k+1)) =
	\begin{cases}
	\frac{q^{i-1}}{1+q+\ldots+q^{M-1}} &   \text{  if } \pi_{k+1}(k+1) \in \{y_a\}_{a=1}^{\overline{x_1}-1}, \\
	\frac{q^{i-j-1}}{1+q+\ldots+q^{M-1}} &   \text{  if } \pi_{k+1}(k+1) \in \{y_a\}_{a = \overline{x_j}+1}^{\overline{x_{j+1}}-1} \text { for } j \in \{1,\ldots,i-1\}, \\
	1 &   \text{  if } \pi_{k+1}(k+1)=y_{\overline{x_i}}, \\
	\frac{q^{M+i-j-1}}{1+q+\ldots+q^{M-1}} &   \text{  if } \pi_{k+1}(k+1) \in \{y_a\}_{a = \overline{x_j}+1}^{\overline{x_{j+1}}-1} \text { for } j \in \{i,\ldots,M-1\}, \\
	\frac{q^{i-1}}{1+q+\ldots+q^{M-1}} &   \text{  if } \pi_{k+1}(k+1) \in \{y_a\}_{a=\overline{x_M}+1}^{n-k}, \\
	0 &   \text{  if } \pi_{k+1}(k+1)=y_{\overline{x_j}} \text{ for } x_j \in S \setminus \{x_i\}.
	\end{cases}
	\end{align*}

	The first two cases above correspond to the second case in the induction hypothesis, the $4^{th},5^{th}$ cases above correspond to the first case in the induction hypothesis, the $3^{rd}, 6^{th}$ cases above correspond to the $3^{rd}, 4^{th}$ cases in the induction hypothesis, respectively.
	
	By Lemma \ref{lma:circle}, we know $Pr(\pi_{k} \oplus y_j |\pi_{k-1} \oplus z)$. Finally, we can get 
	$$
	Q = \begin{cases}
	\tfrac{q^{M-p(z|S)+i-1}}{1+q+\ldots+q^{M-1}}& \text{  if } z>x_i \text{ and } z \not\in S , \vspace{0.3 cm}\\ 
	\tfrac{q^{i-p(z|S)-1}}{1+q+\ldots+q^{M-1}} &   \text{  if } z<x_i \text{ and } z \not\in S .
	\end{cases}  
	$$
	Hence our statement holds for $k = K$, too, thus finishing the proof. 
\end{proofref}

\subsection{Putting Things Together for  \cref{thm:choice probability}}

\begin{proofref}{\cref{thm:choice probability}}
	Name all items as $\{y_z\}_{z=1}^{n}$ such that $y_1 < y_2 < \ldots < y_{n}$. We have
	\begin{equation*}
	Pr(\{x_i|S\}) = 
	\sum_{z=1}^{n} Pr(\{x_i|S\}| (y_z)) \cdot \lambda( (y_z) ) 
	\end{equation*} 
	Use $\overline{x_j}$ to denote the position of item $x_j$ in the universe, hence item $x_j$ is renamed as $y_{\overline{x_j}}$ under ``y" notation. By Lemma \ref{lma:oam_closed_form}, we have the following equation:
	$$
	Pr(\{x_i|S\}| (y_z)) =
	\begin{cases}
	\frac{q^{i-1}}{1+q+\ldots+q^{M-1}} &   \text{  if } y_z \in \{y_a\}_{a=1}^{\overline{x_1}-1}, \\
	\frac{q^{i-j-1}}{1+q+\ldots+q^{M-1}} &   \text{  if } y_z \in \{y_a\}_{a = \overline{x_j}+1}^{\overline{x_{j+1}}-1} \text { for } j \in \{1,\ldots,i-1\}, \\
	1 &   \text{  if } y_z = y_{\overline{x_i}}, \\
	\frac{q^{M+i-j-1}}{1+q+\ldots+q^{M-1}} &   \text{  if } y_z \in \{y_a\}_{a = \overline{x_j}+1}^{\overline{x_{j+1}}-1} \text { for } j \in \{i,\ldots,M-1\}, \\
	\frac{q^{i-1}}{1+q+\ldots+q^{M-1}} &   \text{  if } y_z \in \{y_a\}_{a=\overline{x_M}+1}^{n}, \\
	0 &   \text{  if } y_z=y_{\overline{x_j}} \text{ for } x_j \in S \setminus \{x_i\}.
	\end{cases}
	$$
	By Lemma \ref{lma:topk_pmf}, we have $\lambda( (y_z) ) = \frac{q^{z-1}}{1+\ldots+q^{n-1}}$.	After calculation, we get $Pr(\{x_i|S\})=\frac{q^{i-1}}{1+q+\ldots+q^{M-1}}$.
\end{proofref}

\section{Consistency of MLE for OAM (Theorem \ref{thm:oam_consistent})}\label{sec:proof of oam_consistent}

\subsection{Main Body of the Proof}

\begin{proofref}{\cref{thm:oam_consistent}}
	First fix the underlying parameters of the RMJ-based ranking model $ (\pi^\ast, q^\ast) $. Let the choice data $ H_T = (S_1, x_1, \ldots, S_T, x_T) $ be given, where $ S_t $ is the display set shown to the $ t $th participant and $ x_t $ is his/her choice. Invoking Proposition 3 of \cite{feng2021robust}, the MLE problem for OAM, the choice model induced by the RMJ-based ranking distribution, can be written as
	\begin{align*}
	(\hat{\pi}, \hat{q}) \in
	\, \text{argmax}_{\pi, q} \quad  \sum_{t=1}^{T} \log \left(\frac{1-q}{1 - q^{|S_t|}}\right) + \log q \sum_{(i,j): i \neq j} \I{j \succ_\pi i} w_{ij}
	\label{eq:oam_mle_full},
	\end{align*}
	where
	$$w_{i j} :=\sum_{t=1}^{T} \mathbb{I} \Big\{\{i, j\} \subseteq S_{t} \text{ and }x_{t}=i\Big\}$$
	is the number of times that both items $ i $ and $ j $ are displayed and item $ i $ is chosen (among the $ T $ samples). Furthermore, recall that $ \hat{\pi} $ can be obtained from the integer programming formulation \eqref{eqn:choiceAgg}, which is further equivalent to the following formulation by substituting the relation $ x_{ij} + x_{ji} = 1 $:
	\begin{equation}\label{eqn:choiceAgg2}
	\begin{aligned}
	\{\hat{x}_{ij}: i<j\} \ \in \ \text{argmin}_{\{x_{ij}: i<j\}} \quad &\ssum_{(i,j): i < j} \  (w_{ji} - w_{ij})  \, x_{ij}\\
	\textrm{s.t.} \quad & \begin{array}{lll}
	x_{i j}+x_{j r} - x_{ir} \leq 1 & \quad &\forall\  1 \leq i <  j < r \leq n \\
	x_{i j} \in\{0,1\} & \quad &\forall \ 1 \leq i <  j \leq n
	\end{array}
	\end{aligned}
	\end{equation}
	
	The ranking $ \hat{\pi} $ is obtained by letting $ i \succ_\pi j $ if and only if $ \hat{x}_{ij}  =1 $. Given the solution of $ \hat{\pi} $, the estimator $ \hat{q} $ is obtained by the one-dimensional convex optimization
	\begin{align}\label{eq:MLE for alpha}
	\hat{\alpha} \in \text{argmin}_{\alpha \in (0, +\infty)} \quad L_T(\alpha):= \alpha  \ssum_{t=1}^{T} \ssum_{x \in S_t} \I{x \succ_{\hat{\pi}} x_t} + \sum_{t=1}^{T} \log {\sum\nolimits_{j=0}^{|S_t|-1} e^{-j\alpha} }
	\end{align}
	so that  $ \hat{\alpha} = - \log \hat{q}  $.
	
	We break the rest of the proof into two parts: the ``if" part and the ``only if" part.
	\vspace{0.5 cm}
	
	\underline{The ``if" part.}  Let $ \calS_{\infty} $ be the collection of display sets that are displayed infinite times. Suppose for every pair of items $ \{i,j\} $ is covered infinitely many times. That is, there exists a display set $ S \in \calS_{\infty}  $ such that $ \{i,j\} \subseteq S $. We wish to show that $ (\hat{\pi}, \hat{q}) \to (\pi^\ast, q^\ast) $ almost surely as the sample size $ T \to \infty $.
	
	We first claim that $ \hat{\pi} \to \pi^\ast $ almost surely. Without loss of generality, assume $ \pi^\ast = e $, the identity ranking. In other words, we wish to show that with probability one, there exists  $ \tau $ such that for all $ T \geq \tau $,  the unique solution to \eqref{eqn:choiceAgg2} is $ \{x_{ij} = 1, i < j\} $. Pick an arbitrary pair of items $ i, j $ such that $ i < j $. Let $ N_{ij} := \sum_{t=1}^{T} \mathbb{I} \big\{\{i, j\} \subseteq S_{t} \big\} $ be the number of times that both items $ i $ and $ j $ are displayed. Because both $ i $ and $ j $ are covered by some $ S \in \calS_{\infty} $, we have $ N_{ij} \to \infty$ as $ T \to \infty $. Note that invoking the choice probabilities in \eqref{eqn:oamProb}, OAM is a $ q^\ast $-separable choice model, i.e., for every display set $ S $ such that $ \{i,j\} \subseteq S $, $ \Pr(\{j|S\}) \leq  q \Pr(\{i|S\})$; see \cite{feng2021robust} for more details. Since the choices are generated independently conditional the display sets, we invoke the law of large numbers and conclude that $ w_{ij} \to \infty $, $ w_{ji} \to \infty $, and $ w_{ji} / w_{ij} \to q' $ for some $ q' \leq q $ as the sample size $ T \to \infty $. Furthermore, since there are only a finite number of pairs, with probability one, there exists $ \tau $ such that for all $ T \geq \tau $, 
	\begin{align*}
	w_{ji} - w_{ij} <  0\quad \text{for all } i<j.
	\end{align*}
	In that case, it is straightforward to see that the unique solution to \eqref{eqn:choiceAgg2} is $ \{x_{ij} = 1, i < j\} $. The ``if" part is hence completed by noting the result below. Its proof resembles the standard argument for consistency of MLE except for a few technical differences, such as allowing for an arbitrary display set offering process (which leads to not necessarily i.i.d choice data) and non-compactness of the range of $ \alpha $. We provide the proof details in \cref{sec:proof of consistency of q}.
	\begin{lemma}\label{lma:consistency of q}
		$ \hat{q} \to q^\ast $ almost surely.
	\end{lemma}

	\vspace{0.5 cm}
	
	\underline{The ``only if" part.} Suppose there exists a pair of items $ \{i,j\} $ that is not covered infinitely many times. We wish to show that for some underlying parameter $ (\pi^\ast, q^\ast) $, $ (\hat{\pi}, \hat{q}) \not \to (\pi^\ast, q^\ast) $ with positive probability as the sample size $ T \to \infty $.
	
	Through a relabeling argument, assume $ i = n-1 $ and $ j = n $ without loss of generality. That is, the items $ \{n-1,n\} $ are only displayed together finitely many times. Suppose $ \pi^\ast = e $ is the ground truth ranking. We claim that 
	it does not hold that  $\hat{\pi} \to e $ almost surely as $ T \to \infty $. 
	
	The rest of proof consists of two steps. First, note that the items $ \{n-1,n\} $ are only displayed finitely many times. Therefore, with a positive probability, there exists $ \bar{T} >0 $ such that $ w_{n-1,n} \leq w_{n, n-1} $ for all $ T \geq \bar{T} $. (In particular, if $ \{n-1,n\} $ is not covered \textit{at all}, we have $ w_{n-1,n} = w_{n, n-1} = 0 $ for all $ T \geq 1 $.) 
	
	Second, for the sake of contradiction, suppose $\hat{\pi} \to e $ almost surely as $ T \to \infty $. Then with probability one, there exists $ \tau < + \infty $ so that for all $ T \geq \tau $, $ \hat{\pi} = e $. It further implies that $ \hat{x} := \{\hat{x}_{ij} = 1, \text{ for all } i<j\} $ is an optimal solution to \eqref{eqn:choiceAgg2}. However, notice that if $ w_{n-1, n} \leq w_{n, n-1} $, the solution $ \tilde{x} $ defined by
	\begin{align*}
	\tilde{x}_{ij} = 
	\begin{cases}
	1, &\quad \text{ if } i<j \text{ and }  (i,j) \neq (n-1, n) \\
	0, &\quad \text{ if } i = n-1 \text{ and }  j = n
	\end{cases} 
	\end{align*}
	weakly decreases the objective function value. It is also feasible because $ \tilde{x} $ corresponds to the ranking $ \tpi := (1, 2, \ldots, n-2, n, n-1) $ (i.e., the ranking obtained by swapping the rankings of the bottom two ranked items of $ e $). Therefore, $ \tilde{x} $ is also an optimal solution to \eqref{eqn:choiceAgg2}. Invoking the first step, we know that with positive probability, both $ e $ and $ \tpi $ are MLEs for all $ T \geq \max\{\tau, \bar{T}\} $.
	Note that the tie-breaking rule for MLE cannot be item-specific (for otherwise $ \tpi $ cannot be correctly identified if it is the ground truth ranking). Therefore, with positive probability, no item-blind tie-breaking rule for MLE can differentiate between items $n-1$ and $n$ and thus between rankings $ e $ and $ \tilde{\pi} $. As a result, we conclude that it cannot hold that $\hat{\pi} \to e $ almost surely as $ T \to \infty $.
\end{proofref}

\subsection{Proof of Auxiliary Results (Lemma \ref{lma:consistency of q})}\label{sec:proof of consistency of q}

\begin{proofref}{\cref{lma:consistency of q}}
	Let $ \alpha^\ast := - \log q^\ast \in (0, +\infty) $ and recall that $ \hat{\alpha} $ is solution to \eqref{eq:MLE for alpha}. It suffices to show that $ \hat{\alpha} \to \alpha^\ast $ almost surely. We employ a pathwise analysis throughout the proof.
	
	Given display set $ S $ and item $ x \in S $, let $ N_{S} := \sum_{t=1}^{T} \mathbb{I} \big\{S_{t} = S  \big\} $ and $ N_{S}^x := \sum_{t=1}^{T} \mathbb{I} \big\{S_{t} = S \text{ and } x_t = x \big\} $ be the number of times that $ S $ is displayed and $ x $ is chosen out of $ S $, respectively. For $ i \in \{1, \ldots, |S|\} $, let $ \pi_S(i) $ be the $ i^{th} $ most preferred item within display set under ranking $ \pi^\ast $. In other words, if $ \pi_S(i) = y $, then $ \sum_{x \in S} \I{x \succ_{\pi^\ast} y} = i-1$. Finally, let us introduce
	\begin{align*}
	L_T^S(\alpha) := &\frac{1}{N_S}\Big (\alpha  \sum_{t: S_t = S} \sum_{x \in S} \I{x \succ_{\pi^\ast} x_t} + \sum_{t: S_t = S} \log {\sum\nolimits_{j=0}^{|S|-1} e^{-j\alpha} } \Big)\\
	= & \alpha \sum_{i=1}^{|S|} \frac{N_S^{\pi_S(i)}}{N_S} (i-1) + \log {\sum\nolimits_{j=0}^{|S|-1} e^{-j\alpha} }
	\end{align*}
	to be the (scaled) partial log likelihood loss function when $ \hat{\pi} = \pi^\ast $ and only display set $ S $ is considered. Since the choices are generated independently conditional the display sets, we invoke the choice probabilities in \eqref{eqn:oamProb} as well as the law of large numbers and conclude that with probability one, for all $ S \in \calS_{\infty} $ and $ \alpha \in (0, +\infty) $,
	\begin{align*}
	L_T^S(\alpha) \to L_\infty^S(\alpha):= &\alpha \, \sum_{i=1}^{|S|} \frac{e^{-(i-1)\alpha^\ast}}{\sum\nolimits_{j=0}^{|S|-1} e^{-j\alpha^\ast}} (i-1) + \log {\sum\nolimits_{j=0}^{|S|-1} e^{-j\alpha} } \\
	= &\alpha \, \frac{\sum_{i=1}^{|S|-1}ie^{-i\alpha^\ast}}{\sum\nolimits_{j=0}^{|S|-1} e^{-j\alpha^\ast}}  + \log {\sum\nolimits_{j=0}^{|S|-1} e^{-j\alpha} }.
	\end{align*}
	In fact, the convergence is also locally uniform: for all $ M > 0 $, 
	\begin{align*}
	\sup_{\alpha \in (0, M]}|L_T^S(\alpha) - L_\infty^S(\alpha)| = &\sup_{\alpha \in (0, M]} \alpha \sum_{i=1}^{|S|} \Bigg \vert \frac{N_S^{\pi_S(i)}}{N_S} -  \frac{e^{-(i-1)\alpha^\ast}}{\sum_{j=0}^{|S|-1} e^{-j\alpha^\ast}}\Bigg \vert  (i-1) \\
	\leq &M\sum_{i=1}^{|S|} \Bigg \vert \frac{N_S^{\pi_S(i)}}{N_S} -  \frac{e^{-(i-1)\alpha^\ast}}{\sum_{j=0}^{|S|-1} e^{-j\alpha^\ast}}\Bigg \vert  (i-1) \to 0.
	\end{align*}
	as sample size $ T \to \infty $. Therefore, with probability one,  $ 
	L_T^S(\cdot) \to  
	L_\infty^S(\cdot) $ uniformly on the set $(0, M] $.
	
	In addition, it is straightforward to verify that for every $ S $, $ L^S_\infty(\cdot) $ is strictly convex and attains its unique minimum at $ \alpha^\ast $. In fact, one can verify that the first order condition is
	\begin{align*}	(L_\infty^S)'(\alpha) = 0 \Rightarrow \frac{\sum_{i=1}^{|S|-1}ie^{-i\alpha^\ast}}{\sum\nolimits_{j=0}^{|S|-1} e^{-j\alpha^\ast}}  = \frac{\sum_{i=1}^{|S|-1} ie^{-i\alpha}}{\sum_{j=0}^{|S|-1} e^{-j\alpha}},
	\end{align*}
	which implies that $ \text{argmin}_{\alpha} L_\infty^S(\alpha) = \alpha^\ast $. Since $ 
	L_T^S(\cdot) \to  
	L_\infty^S(\cdot) $ uniformly on the set $(0, M] $ for every $ M >0 $, we further conclude that with probability one and for an arbitrarily small $ \epsilon > 0 $, there exists $ \tau_2  $ such that for all $ T \geq \tau_2 $, 
	\begin{align*}
	L_T(\alpha) = \sum_{S \in \calS_\infty} N_S L_T^S(\alpha) + O(1)
	\end{align*}
	attains its minimum in $ [\alpha^\ast - \epsilon, \alpha^\ast + \epsilon] $. In other words, $ \hat{\alpha} \in [\alpha^\ast - \epsilon, \alpha^\ast + \epsilon] $.
\end{proofref}

\section{Inconsistency of the Mallows Smoothing Heuristic (\cref{thm:ms_unstable})}\label{sec:proof of thm:ms_unstable}

\subsection{Overview}

The Mallows Smoothing heuristic (\cite{Antoine2021}) consists of a two-step process.

\begin{enumerate}[align=left]
	\item [Step 1:] (``Choice to Ranking") This step takes the empirical choice probabilities $ \{ \hat{\Pr}(x|S)\}$ as input and produces a distribution over rankings $ \hat{\lambda} $ as output. The goal of this step is to find $ \hat{\lambda}  $, under which the aggregated choice probabilities match the empirical choice probabilities. 
	
	Formally, let $ \hat{\calS} $ be the collection of display sets that have appeared in the data. 
	This step corresponds to finding a solution $ \hat{\lambda} $ to the following feasibility problem:\footnote{If the system above is not feasible, then solve a ``soft" version of it by minimizing a loss function of residues; see \cite{Jagabathula2018} for more details.}
	\begin{equation}\label{eq: feasibility problem for MS}
	\begin{aligned}
	\min_{\hat{\lambda} }& \quad 0 \\
	\ssum_{\tpi \in \Sigma}  \hat{\lambda}(\tpi) \, \II{x, \tpi, S} &= \hat{\Pr}(x|S), &\text{ for all } S \in \hat{\calS} \text{ and } x\in S \\
	\sum_{\tpi \in \Sigma} \hat{\lambda} (\tpi) &= 1 \\
	\hat{\lambda} (\pi) & \geq 0, &\text{ for all } \pi \in \Sigma.
	\end{aligned}
	\end{equation}

	\item [Step 2:]  (``Smoothing")
	This step takes a distribution over rankings $ \hat{\lambda} $ as input and produces the Mallows distribution parameters $ (\hat{\pi}, \hat{q}) $ as output. The goal of this step is to  find a Mallows distribution that fits the $ \hat{\lambda} $ distribution produced by the previous step.

	
	Formally, recall that the Kendall's Tau distance between two rankings $ \pi, \tpi $ is given by $d_K(\pi, \tpi) = \sum_{x<y} \mathbb{I}\{\left(\pi^{-1} (x)-\pi^{-1} (y)\right) \cdot\left(\tpi^{-1}(x)-\tpi^{-1}(y)\right)<0\}  $. The estimated central ranking $ \hat{\pi} $ of the Mallows distribution is obtained from the following ranking aggregation problem:
	\begin{align}\label{eq:ranking aggregation}
	\hat{\pi} \in \text{argmin}_{\pi} \ \sum_{\tpi} \hat{\lambda}(\tpi) \cdot d_K(\pi, \tpi).
	\end{align}

	The dispersion parameter is obtained from solving the following convex problem:
	\begin{align*}
	\hat{q} = \text{argmin}_{q}  \ q \cdot \sum_{\tpi} \hat{\lambda}(\tpi) \cdot d_K(\hat{\pi}, \tpi) + T  \log \psi(n, q).
	\end{align*}
	
\end{enumerate}

Note that in system \eqref{eq: feasibility problem for MS}, the number of variables is on the order of $ \Omega(n!) $ while the number of constraints is on the order of $ O(n2^n) $, which is much smaller. Therefore, whenever feasible, the solution forms a high dimensional polytope. 

Roughly speaking, we will show that even when all display sets with sizes at least three are displayed infinite many times, the Mallows model parameters cannot be identified from the MS heuristic. That is, we suppose that infinite choice data is generated from a Mallows model with the central ranking to be $ e $. However, we can construct a ranking distribution $ \tilde{\lambda} $ that solves \eqref{eq: feasibility problem for MS} so that when we use it as input to problem \eqref{eq:ranking aggregation}, we find a solution $ \tilde{\pi} $ that is different from $ e $. That is, a \textit{wrong} model parameter can be the output of the MS heuristic.

\subsection{Main Body of the Proof}
\begin{proofref}{\cref{thm:ms_unstable}}
	Pick $ q \in (0,1) $. Let $  \lambda^e $ be the p.m.f. of the Mallows ranking model with the central ranking to be the identity ranking $ e $ and the dispersion parameter to be $ q$. Let $ \{\Pr^e (x|S): |S| \geq 3, x \in S\} $ be the collection of associated choice probabilities for display sets with sizes at least three, formally defined as $ \Pr^e (x|S) :=  \ssum_{\tpi \in \Sigma}  \lambda^e(\tpi) \, \II{x, \tpi, S}$ for all $ S $ such that $ |S| \geq 3 $ and  $ x \in S $. 
	
	We claim that we can construct $ \tilde{\lambda} \in \Delta(\Sigma) $ that satisfies two properties simultaneously. 
	\begin{enumerate}
		\item First, it solves \eqref{eq: feasibility problem for MS} when taking $ \{\Pr^e (x|S): |S| \geq 3, x \in S\} $ as input. That is, $ \tilde{\lambda} \in \Delta(\Sigma) $ satisfies
		\begin{equation}\label{eq:choice2ranking}
		\ssum_{\tpi \in \Sigma}  \tilde{\lambda}(\tpi) \, \II{x, \tpi, S} = \Pr\nolimits^e (x|S), \text{ for all } S \text{ such that } |S| \geq 3 \text{ and } x\in S.
		\end{equation}
		\item Second, if we take $ \tilde{\lambda} $ as input to \eqref{eq:ranking aggregation}, the (wrong) ranking $ \tilde{\pi} := (1, 2, \ldots, n-2, n, n-1) \neq e $ can be obtained as solution. That is, 
		\begin{align}\label{eq:wrong ranking}
		(1, 2, \ldots, n-2, n, n-1) \in \text{argmin}_{\pi} \sum_{\tpi} \tilde{\lambda}(\tpi) \cdot d_K(\pi, \tpi).
		\end{align}
	\end{enumerate}
	
Our construction is based on two observations. First, we invoke Theorem 3.7 by \cite{Jagabathula2018} and know that under the current display set setup, \eqref{eq: feasibility problem for MS} can only identify a ranking up to its first $ n-2 $ positions. As a consequence, as long as $ \tilde{\lambda} $ satisfies
\begin{align}\label{eq:condition 1}
\sum_{\tpi \in \Sigma} \tilde{\lambda} (\tpi) \, \I{\pi_{n-2} \subseteq \tpi} = \sum_{\tpi \in \Sigma} \lambda^e (\tpi) \, \I{\pi_{n-2} \subseteq \tpi}, \quad \text{ for all } \pi_{n-2} \in \Sigma_{n-2},
\end{align}
it also satisfies \eqref{eq:choice2ranking}.

Second, we observe that  problem \eqref{eq:ranking aggregation} only depends on the \textit{pairwise} choice probabilities associated with its input ranking distribution. This observation is formalized as the result below, and we present its proof in \cref{sec:proof of mleRanking_2}.

\begin{lemma} \label{mleRanking_2}
	Pick $ \hat{\lambda} \in \Delta(\Sigma) $ and let $ P_{x,\{x,y\}} := \ssum_{\tpi}   \hat{\lambda}(\tpi) \I{ \tpi^{-1}(x) < \tpi^{-1}(y)}$ be its associated pairwise probability of choosing item $ x $ out of $ \{x,y\} $. Then 
	\begin{equation*}
	\operatorname*{argmin}_{\pi} \sum_{\tpi}  \hat{\lambda}(\tpi) \cdot d_K(\pi, \tpi) = \operatorname*{argmin}_\pi \sum_{x<y} \I{\pi^{-1} (x) <\pi^{-1} (y)} \cdot (1-2P_{x,\{x,y\}}).
	\end{equation*}
\end{lemma}
Let $\tilde{P}_{x,\{x,y\}}:= \ssum_{\tpi}  \tilde{\lambda}(\tpi) \I{ \tpi^{-1}(x) < \tpi^{-1}(y)}$
As a quick consequence of the result above, a sufficient condition of \eqref{eq:wrong ranking} is
\begin{equation}\label{eq:condition pairse}
\tilde{P}_{x,\{x,y\}}>1/2 \text{ for all } (x,y) \in \{(x,y): x < y, (x,y) \neq (n-1,n)\} \text{ and } \tilde{P}_{n-1,\{n-1,n\}}<1/2.
\end{equation}

In order to construct $ \tilde{\lambda} $ that satisfies both \eqref{eq:condition 1} and \eqref{eq:condition pairse} (and therefore, fulfills out claim), let us classify $ \Sigma $ into 3 groups based on its top-$(n-2)$ elements. Specifically, given $ \pi \in \Sigma $ and $ \pi_{n-2} \in \Sigma_{n-2} $ such that $ \pi_{n-2} \subseteq \pi $, we categorize $ \pi $ into three of the groups below:
\begin{enumerate}[align = left]
	\item [Group 1: ]  Under $ \pi_{n-2} $, item $ (n-1) $ is preferred to item $ n $ . That is, either
	\begin{enumerate}
	    \item [(i)] $ \{n-1, n\} \subseteq R(\pi_{n-2}) $ and $ \pi_{n-2}^{-1}(n-1)<\pi_{n-2}^{-1}(n) $; or
	    \item [(ii)] $ (n-1) \in R(\pi_{n-2}) $ but $ n \in R^c(\pi_{n-2}) $. 
	\end{enumerate}
	\item [Group 2: ] Under $ \pi_{n-2} $, item $ n $ is preferred to item $ (n -1) $. That is, either 
	\begin{enumerate}
		\item [(i)] $ \{n-1, n\} \subseteq R(\pi_{n-2}) $ and $ \pi_{n-2}^{-1}(n-1)>\pi_{n-2}^{-1}(n) $, or
		\item [(ii)] $ n \in R(\pi_{n-2}) $ but $ (n-1) \in R^c(\pi_{n-2}) $.
	\end{enumerate}

	\item [Group 3: ] Under $ \pi_{n-2} $, items $ n $ and $ (n-1) $ are incomparable. That is $ \{n-1, n\} = R^c (\pi_{n-2}) $. 
\end{enumerate} 

We explicitly construct $ \tilde{\lambda} $ as the following:
\begin{align*}
\tilde{\lambda} (\pi) = 
\begin{cases}
\lambda^e (\pi) & \text{ if } \pi \text{ belongs to Group 1 or 2} \\
0 & \text{ if } \pi \in \text{ Group 3 and } (n-1) \succ_\pi n \\
\lambda^e(\pi_{n-2} \oplus (n-1, n)) + \lambda^e(\pi_{n-2} \oplus (n, n-1) ) & \text{ if } \pi \in \text{ Group 3 and } n \succ_\pi (n -1 )
\end{cases}
\end{align*}
In other words, $ \tilde{\lambda} $ is obtained from $ \lambda^e $ by ``transporting"  weights in favor of item $ n $ over item $ (n-1) $ when both of those items are ranked at the bottom two. Since the top-$ (n-2) $ rankings are not disturbed, \eqref{eq:condition 1} is satisfied by construction. 

The rest of proof is devoted to verifying \eqref{eq:condition pairse}. Let us use $ \lambda = \lambda^e $ and its associated pairwise choice probability $ P_{x,\{x,y\}} $ for shorthand notation. Since only items $ (n-1) $ and $ n $ are swapped when they are ranked at the bottom, we have for all $ (x,y) \in \{(x,y): x < y, (x,y) \neq (n-1,n)\} $,
\begin{align*}
\tilde{P}_{x,\{x,y\}} =  P_{x,\{x,y\}} > 1/2.
\end{align*}
Denote the sum of probabilities (under the Mallows model) of top-$(n-2)$ rankings in Group $1$ as $ \lambda(\text{Group 1}):= \ssum_{\tpi}  \lambda(\tpi) \I{ \tpi \in \text{ Group 1}}$. Given the construction rule, we have $\tilde{P}_{n-1,\{n-1,n\}} =  \lambda(\text{Group 1})$. Therefore, to verify \eqref{eq:condition pairse}, it suffices to show
\begin{align*}
\lambda(\text{Group 1}) := \sum_{\pi \in \text{Group 1}} \lambda^e (\pi) < \tfrac{1}{2}.
\end{align*}
		
In order to show that $ \lambda(\text{Group 1}) < 1/2 $, note that Groups 1, 2, and 3 are defined based on the top-$(n-2)$ items of a ranking. In light of this, let us build on \cite{collas2021concentric}, who characterize the probability distribution of top-$ (n-2) $ rankings under the Mallows model. Formally speaking, we use $\lambda_{n-2}: \Sigma_{n-2} \to \mathbb{R} $ to denote the pmf of a top-($n-2$) ranking under Mallows model. That is, $ \lambda_{n-2} (\pi_{n-2}) := \sum_{\pi} \I{\pi_{n-2} \subseteq \pi} \lambda^e (\pi)$.  Also, in the proof below,  given a collection of top-$ (n-2) $ rankings $ E \subseteq \Sigma_{n-2} $, we use $ \Pr(E) := \sum_{\pi_{n-2} \in E} \lambda_{n-2}(\pi_{n-2}) $ as a shorthand notation to represent the total mass of top-$ (n-2) $ rankings in $ E $. 

We (re-)classify all top-$(n-2)$ rankings depending on the positions of items $ (n-1) $ and $ n $. Given a top-$(n-2)$ ranking $ \pi_{n-2} $, we call an item $ x $ a \textit{head item} if it ranks in top-$(n-2)$, i.e., $ x \in R(\pi_{n-2}) $. Otherwise, we call $ x $ a \textit{tail item}. We describe the classification of top-$ (n-2) $ rankings below.

	\begin{enumerate}
		\item \textit{Class $ A $}: \textit{Both items $ (n-1) $ and item $n$ are tail items.} That is, $ \pi_{n-2} \text{ belongs to } \text{Class $ A $} $ if $ \{n-1, n\} = R^c (\pi_{n-2}) $. 
		We can further partition Class $ A $ into 2 subclasses depending on the two bottom-ranked items of the top-$ (n-2) $ ranking:
		\begin{enumerate}
			\item \textit{Class $A_1$}: $\pi_{n-2} (n-3) <\pi_{n-2} (n-2).$ 
			\item \textit{Class $A_2$}: $\pi_{n-2} (n-3) >\pi_{n-2} (n-2).$ 
		\end{enumerate}
		
		There is a one-to-one correspondence (i.e., bijection) between Classes $ A_1 $ and $ A_2 $. For a top-$(n-2)$ ranking $a_1$ in Class $ A_1 $, we can construct a top-$(n-2)$ ranking $ a_2$ in Class $ A_2 $ by swapping two bottom-ranked items ranked at positions $ (n-3) $ and $ (n-2) $:
		\begin{align*}
        a_2(i) = 
        \begin{cases}
        a_1(i) & \text{ if } i < n-3; \\
        a_1(n-2) & \text{ if } i = n-3; \\
        a_1(n-3) & \text{ if } i = n-2.
        \end{cases}
        \end{align*}
        Before proceeding, let's restate a fact from \cite{collas2021concentric}.
        
       \vskip 0.3 cm
        \textbf{Fact} \textit{ (Restated from Lemma 1 of \cite{collas2021concentric})
            The probability of a top-k ranking $\pi_k$ under the Mallows ranking model with the identity central ranking is
            \begin{align*}
            \lambda_k(\pi_{k})=q^{d_K(\pi_k)} \cdot \frac{\psi(n-k, q)}{\psi(n, q)}, 
            \end{align*}
            where  $d_K(\pi_k) = \ssum_{i=1}^{k} \left(\ssum_{j=i+1}^{k}  \mathbb{I}\{\pi_k(i)>\pi_k(j)\} + \ssum_{j \in R^c(\pi_k)} \mathbb{I}\{\pi_k(i)>j\}\right)$ is the  Kendall's tau distance between top-$ k $ ranking $ \pi_k $ and the identity ranking.
        }
    \vskip 0.3 cm
    
    In particular, if we take $ k = n-2 $ in the result above, we have
    \begin{align}\label{eq:top-k ranking under Mallows}
    \lambda_{n-2}(\pi_{n-2})=q^{d_K(\pi_{n-2})} \cdot \frac{\psi(2, q)}{\psi(n, q)}, 
    \end{align}
    where
    \begin{align*}
    d_K(\pi_{n-2}) = \sum_{i=1}^{n-2} \left(\sum_{j=i+1}^{n-2}  \mathbb{I}\{\pi_{n-2}(i)>\pi_{n-2}(j)\} + \sum_{j \in R^c(\pi_{n-2})} \mathbb{I}\{\pi_{n-2}(i)>j\}\right).
    \end{align*}
    
    For each bijection pair ($a_1, a_2$), $d_K(a_2)=d_K(a_1)+1$. Hence, $\lambda_{n-2}(a_2) = q \cdot \lambda_{n-2}(a_1)$. 
        
        Let $ P_{A}:=\Pr(\text{Class }A)$ and $ P_{A_i}:=\Pr(\text{Class }A_i)$ for $ i\in \{1,2\} $. In light of the bijection construction above, we have $P_{A_2} = q \cdot P_{A_1}$ and $P_{A} = P_{A_1}+P_{A_2}.$ Therefore, we have
        \begin{align}\label{eq:class A}
        P_{A_1} = \frac{1}{1+q}P_{A} \quad \text{ and } \quad P_{A_2} = \frac{q}{1+q}P_{A}.
        \end{align}

		\item \textit{Class $ B $}: \textit{Item $ (n-1) $ is a head item and item $n$ is a tail item.} That is, $ \pi_{n-2} $ belongs to Class $ B $ if item $ n-1 \in R (\pi_{n-2}) $ but item $ n \not\in R (\pi_{n-2}) $. We can further partition Class $ B $ into $ n-2 $ subclasses by the position of item $(n-1)$. For every $ m \in \{1, 2, \ldots, n-2\} $, the top-$ (n-2) $ ranking $ \pi_{n-2} $ belongs to the subclass $ B_m $ if it belongs to Class $ B $ and item $(n-1)$ is the $ m^{th} $ preferred, i.e., $ \pi_{n-2} (m) = n-1 $.
		
		For every $ m \in \{1, 2, \ldots, n-2\} $, there is a bijection between Classes $ A $ and $ B_m $. For any top-$(n-2)$ ranking $a \in \text{Class } A$, we can obtain a top-$(n-2)$ ranking $b_m \in \text{ Class } B_m$ from the following construction:
		\begin{align*}
		b_m(i) = 
		\begin{cases}
		a(i) & \text{ if } i < m; \\
		n-1 & \text{ if } i = m; \\
		a(i-1) & \text{ if } m + 1 \leq i \leq n-2.
		\end{cases}
		\end{align*}
		In other words, $ b_m $ is obtained from $ a $ by first putting item $(n-1)$ at the $m^{th}$ position and, moving the original $m^{th},(m+1)^{th},\ldots,(n-1)^{th}$ items back one position, and finally leaving the original $(n-2)^{th}$ as a tail item. Furthermore, by \eqref{eq:top-k ranking under Mallows}, we have $d_K(b_m)=d_K(a)+n-1-m$ and $\lambda_{n-2}(b_m) = q^{n-1-m} \cdot \lambda_{n-2}(a)$. 
		
		Let $ P_{B}:=\Pr(\text{Class }B)$ and  $ P_{B_m}:=\Pr(\text{Class }B_m)$ for all $ m \in \{1, 2, \ldots, n-1\} $. In light of bijection construction above, we have $P_{Bm} = q^{n-1-m} \cdot P_{A}$ and $ P_B = \sum_{m=1}^{n-2} P_{Bm}  $. Therefore,
		\begin{align}\label{eq:class B}
		P_{B} = \sum_{m=1}^{n-2} q^{n-1-m} \cdot P_{A}.
		\end{align}

		\item \textit{Class $ C $}: \textit{Item $n$ is a head item and item $ (n-1) $ is a tail item.}  That is, $ \pi_{n-2} $ belongs to Class $ C $ if item $ n-1 \not \in R (\pi_{n-2}) $ but item $ n \in R (\pi_{n-2}) $.
		
		There is a bijection between Classes $B$ and $C$. For any top-$(n-2)$ ranking $b \in \text{Class }B$,  we can obtain a top-$(n-2)$ ranking $c \in \text{Class }C$ from the following construction:
		\begin{align*}
		c(i) = 
		\begin{cases}
		b(i) & \text{ if } i \neq b^{-1}(n-1); \\
		n & \text{ if } i = b^{-1}(n-1).
		\end{cases}
		\end{align*}
		In other words, $c$ is obtained from $b$ by replacing item $(n-1)$ by item $n$. 
        By \eqref{eq:top-k ranking under Mallows}, we have $d_K(c)=d_K(b)+1$ and $\lambda_{n-2}(c) = q \cdot \lambda_{n-2}(b)$.
		Let $ P_{C}:=\Pr(\text{Class }C)$, and we thus have 
		\begin{align}\label{eq:class C}
		P_{C} = q   P_{B} = \sum_{m=1}^{n-2} q^{n-m} \cdot P_{A}.
		\end{align}

		\item \textit{Class $ D $}: \textit{Item $ (n-1) $ and item $n$ are both head items.} That is, $ \pi_{n-2} $ belongs to Class $ D $ if  $ \{n-1, n\} \subseteq R(\pi_{n-2})$. We can partition Class $ D $ into $(n-2)(n-3)$ subclasses based on the preference positions of items $(n-1)$ and $n$. More specifically, given $ j \neq k $ and $j , k \in \{1, \ldots, n-2\}$, let the subclass $$  D_{(j,k)}:= \{\pi_{n-2} \in \text{Class } D: \pi_{n-2}(j)=n-1 \text{ and }\pi_{n-2}(k) = n\}$$ be the collection of top-$(n-2)$ rankings  such that items $(n-1)$ and $n$ are ranked in the $j^{th}$ and $k^{th}$ positions, respectively. Furthermore, let the subclass
		\begin{align*}
		D_{1} := \cup_{j<k} \text{ Subclass } D_{(j,k)} \quad (\text{resp. }\quad D_{2} := \cup_{j>k} \text{ Subclass } D_{(j,k)} ) 
		\end{align*}
		be the collection of top-$ (n-2) $ rankings that rank item $ (n-1) $ higher (resp. lower) than item $ n $.
		
		Given $j < k$, there is a bijection between subclass $ D_{(j,k)} $ and Class $A_1$. For any ranking $a_1 \in \text{Class }A_1$, we can obtain a top-$(n-2)$ ranking $d_{{(j,k)}} \in \text{Subclass } D_{(j,k)}$ from the following construction:
		\begin{align*}
		d_{{(j,k)}}(i) = 
		\begin{cases}
		a_1(i) & \text{ if } i < j \\
		n-1 & \text{ if } i = j \\
		a_1(i-1) & \text{ if } j < i < k \\
		n & \text{ if } i = k \\
		a_1(i-2) & \text{ if } i > k 
		\end{cases}
		\end{align*}
		In other words,	to obtain $ d_{(j,k)} $ from $ a_1 $, we can put items $(n-1)$ and $ n $ at positions $j$ and $k$ respectively, move the original $j^{th},(j+1)^{th},\ldots, (k-2)^{th}$ items back one position, move the original $(k-1)^{th},\dots,(n-4)^{th}$ items back two positions, and finally leave $ \{ a_1(n-3), a_1(n-2)\} $ as tail items. By \eqref{eq:top-k ranking under Mallows}, we have $d_K(d_{{(j,k)}})=d_K(a_1)+2n-1-j-k$ and $\lambda_{n-2}(d_{{(j,k)}}) = q^{2n-1-j-k} \cdot \lambda_{n-2}(a_1)$.  Let $ P_{D_{(j,k)}}:=\Pr(\text{Class }D_{(j,k)})$, and we thus have 
		\begin{align}\label{eq:class Djk}
		P_{D_{(j,k)}} = q^{2n-1-j-k} \cdot P_{A_1}, \quad \text{ for every } j < k.
		\end{align}
		
		There is also a bijection between subclass $ D_{1} $ and subclass $ D_{2} $. This bijection is similar to that of Classes $ A_1 $ and $ A_2 $. For every $ d_{1} \in \text{Subclass } D_{1} $, we can swap the positions of items $n$ and $(n-1)$ to obtain a top-$(n-2)$ ranking $d_{2} \in \text{Subclass } D_{2}$. By \eqref{eq:top-k ranking under Mallows}, we have $d_K(d_{2})=d_K(d_{1})+1$ and $\lambda_{n-2}(d_{2}) = q \cdot \lambda_{n-2}(d_{1})$. Let $ P_{D_i}:=\Pr(\text{Class }D_i)$ for $ i = 1,2 $, and we thus have
		\begin{align}\label{eq:class D2}
		P_{D_2} = q \cdot P_{D_1}.
		\end{align}
		Furthermore, let $ P_{D}:=\Pr(\text{Class }D)$.  Invoking \cref{eq:class A,eq:class Djk,eq:class D2}, we have
		\begin{align}\label{eq:class D}
		\begin{split}
		P_{D} &= P_{D_1} + P_{D_2} =  (1+q) \cdot P_{D_1} = (1+q) \cdot \ssum_{j=1}^{n-3} \ssum_{k=j+1}^{n-2} P_{D_{(j,k)}} \\
		&= (1+q) \cdot \ssum_{j=1}^{n-3} \ssum_{k=j+1}^{n-2} q^{2n-1-j-k} \cdot P_{A_1}\\
		&= \ssum_{j=1}^{n-3} \ssum_{k=j+1}^{n-2} q^{2n-1-j-k} \cdot P_{A}
		\end{split}
		\end{align} 
	\end{enumerate}

	Because $P_{A}+P_{B}+P_{C}+P_{D}=1$, we align \cref{eq:class A,eq:class B,eq:class C,eq:class D}, and conclude that $P_{A} = \frac{(1-q)(1-q^2)}{(1-q^{n-1})(1-q^{n})}$.  Therefore, we can obtain the probability of each aforementioned (sub)class.
	

Clearly, there is a correspondence between Groups 1-3 and Classes $ A $-$ D $: Class $ B $ corresponds Group 1 (ii), Class $ D1 $ corresponds Group 1 (i). Class $ C $ corresponds Group 2 (ii), Class $ D2 $ corresponds to Group 2 (i), and finally, Class $ A $ corresponds Group 3. Therefore, 
$$\lambda(\text{Group 1}) = P_{B} + P_{D1} =   \left(1-\frac{(1-q)(1-q^2)}{(1-q^{n-1})(1-q^{n})} \right)\frac{1}{1+q}.$$
The last equality is due to the following chain of argument: $P_{C} = q \cdot P_{B}$ and $P_{D2} = q \cdot P_{D1}$; see \cref{eq:class C,eq:class D2}. Therefore, $P_{A}+P_{B}+P_{C}+P_{D}=P_{A}+P_{B}+P_{C}+P_{D1}+P_{D2}=P_{A}+(1+q)(P_{B} + P_{D1})= 1$.

As a further consequence, $ \lambda(\text{Group 1}) < 1/2$ when $n$ and $q$ satisfy 
	$$ \left(1-\frac{(1-q)(1-q^2)}{(1-q^{n-1})(1-q^{n})} \right)\frac{1}{1+q} < \frac{1}{2},$$ 
	which simplifies to
	\begin{align*}
	\mathcal{F}_n (q) := 1+q^{n-1}+q^{n}-2q^2-q^{2n-1}>0.
	\end{align*}

	Note that $ \mathcal{F}_n (0) = 1 > 0, \mathcal{F}_n (1) = 0, $ and $ \mathcal{F}_n' (1) = (n-1) + n - 4 - (2n-1) < 0 $. Therefore, we conclude that for all $ n $, $ \mathcal{F}_n (q) > 0 $ either when $ q $ is sufficiently close to $ 0 $ or $ 1 $. That finishes the proof.
\end{proofref}

\subsection{Proof of Auxiliary Results (\cref{mleRanking_2})}\label{sec:proof of mleRanking_2}

\begin{proofref}{\cref{mleRanking_2}}
	\begin{gather*}
	\begin{aligned}
	&\text{argmin}_{\pi} \sum_{\tpi} \hat{\lambda}(\tpi) \cdot d_K(\pi, \tpi) \\
	= &\text{argmin}_\pi \sum_{\tpi} \hat{\lambda}(\tpi) \sum_{x<y} \mathbb{I}\{\left(\pi^{-1} (x)-\pi^{-1} (y)\right) \cdot\left(\tpi^{-1}(x)-\tpi^{-1}(y)\right)<0\} \\
	= &\text{argmin}_\pi \sum_{\tpi} \hat{\lambda}(\tpi) \sum_{x<y} [ \mathbb{I} \{\tpi^{-1}(x) <\tpi^{-1}(y)\} + 
	\mathbb{I} \{\pi^{-1}(x) <\pi^{-1}(y)\} \cdot (1-2\mathbb{I} \{\tpi^{-1}(x) <\tpi^{-1}(y)\})] \\ 
	= &\text{argmin}_\pi \sum_{x<y} \mathbb{I} \{\pi^{-1} (x) <\pi^{-1} (y)\} \cdot \sum_{\tpi}  \hat{\lambda}(\tpi) (1-2\mathbb{I} \{\tpi^{-1} (x) <\tpi^{-1}(y)\}) \\
	= &\text{argmin}_\pi \sum_{x<y} \mathbb{I} \{\pi^{-1} (x) <\pi^{-1} (y)\} \cdot (1-2\sum_{\tpi}  \hat{\lambda}(\tpi) \mathbb{I} \{\tpi^{-1} (x) <\tpi^{-1}(y)\}) \\
	= &\text{argmin}_\pi \sum_{x<y} \mathbb{I} \{\pi^{-1} (x) <\pi^{-1} (y)\} \cdot (1-2P_{x,\{x,y\}})
	\end{aligned}
	\end{gather*}
\end{proofref}

\section{On the Ranked Choice Probabilities and Model Estimation for $ k \geq 1 $ (Theorems \ref{thm:multiple choice probability} and \ref{thm:ranked_choice_estimation})}

\begin{proofref}{\cref{thm:multiple choice probability}}
	Without loss of generality, suppose the display set is given by $ S = \{x_1, \ldots, x_M\} $ such that $ x_1 < x_2 < \cdots < x_M $ for some $ M \geq 2 $. We prove this theorem by forward induction on $k$ (i.e., the length of the ranked list). 
	
	\emph{Base step.} Suppose $k=1$. Given $ \pi_1 = (z) $ where $ z \in S $, we have $ d_S(\pi_{1}) = 0$, $ L_S(\pi_{1}) = |\{x \in S \setminus \{z\} : x < z\}| $ is how $ z $ is ranked relatively in the display set $ S $ under the identity ranking $ e $, and $ \tfrac{\psi (|S|-1,q)}{\psi (|S|,q)} = \frac{1}{1 + q +\cdots + q^{|S|-1}}$. Therefore,
	\begin{align*}
	q^{d_S(\pi_{1})+L_S(\pi_{1})} \cdot \frac{\psi (|S|-1,q)}{\psi (|S|,q)}
	\end{align*}
	equals the choice probability \eqref{eqn:oamProb} in Theorem \ref{thm:choice probability}.

	\emph{Inductive step.} Pick an arbitrary $K \in\{2, \ldots, M\}$. Suppose our statement holds for every $k=1, \ldots, K-1$. We want to show that our statement holds for $k=K$.
	
	Pick an arbitrary top-$ k $ ranking $\pi_k = (x_1^k,x_2^k,\ldots,x_{k-1}^{k},x_k^k) $ and display set $ S $ such that $ R(\pi_k) \subseteq S $. For shorthand notation, let $\pi_{k-2} := (x_1^k,x_2^k,\ldots,x_{k-2}^{k} )$ and $\pi_{k-1} := (x_1^k,x_2^k,\ldots,x_{k-1}^{k} )$ so that $ \pi_{k-1} = \pi_{k-2} \oplus x_{k-1}^{k} $ and $ \pi_{k} = \pi_{k-1} \oplus x_{k}^{k} $. Note that $ \{\pi_k|S\} $ (resp. $ \{\pi_{k-1}|S\} $) is the event that the ranked list $ \pi_k $ (resp. $ \pi_{k-1} $) belongs to the top $ k $ (resp. $ k-1 $) most preferred items within the display set $ S $. Therefore, we have
	\begin{equation} \label{eqn:repeated selection}
	\Pr(\{\pi_k|S\}) = \Pr\big(\{\pi_{k-1}|S\}\big) \cdot \Pr\Big( \big\{x_k^k \,|\,S \setminus R(\pi_{k-1}) \big\} \,|\, \{\pi_{k-1}|S\}\Big). 
	\end{equation}

	There are two parts of RHS of \eqref{eqn:repeated selection}. We first evaluate the first part: by the induction hypothesis, 
	\begin{align} \label{eq:first_part_RHS}
	\Pr\big(\{\pi_{k-1}|S\}\big) = q^{d_S(\pi_{k-1})+L_S(\pi_{k-1})} \cdot \frac{\psi (|S|-k+1,q)}{\psi (|S|,q)}.
	\end{align}
	Now we analyze the second part, which is the probability that $x_k^k$ ranks the first among $S \setminus R(\pi_{k-1})$, given that the ranked list $\pi_{k-1}$ ranks the top-$(k-1)$ among $S$. Given any $m \in \{k-1, k, \ldots, k-1 + n - M\}$, let $\Sigma_{m-1}' \subseteq \Sigma_{m-1}$ be the collection of top-$ (m-1) $ ranked lists so that $ \pi_{m-1}' \in \Sigma_{m-1}' $ if and only if it satisfies the following two properties:
	
	\begin{enumerate}
		\item $R(\pi_{m-1}') \cap S =R(\pi_{k-2})$. That is, items included in $ \pi_{m-1}' $ is either included in $ \pi_{k-2} $, too,  or outside of the display set $ S $.
		\item $\pi_{k-2}$ ranks top-$(k-2)$ among $S$ under $\pi_{m-1}'$;
	\end{enumerate}

	For some quantity $ c $ (to be explained later), we can decompose the second part of RHS of \eqref{eqn:repeated selection} as follows:
	\begin{align*}
	&\phantom{=}\Pr\Big( \big\{x_k^k \,|\,S \setminus R(\pi_{k-1}) \big\} \,|\, \{\pi_{k-1}|S\}\Big) \\
	&=\sum_{m=k-1}^{n-M+k-1}  \sum_{\substack{\pi_{m-1}' \in \Sigma_{m-1}^{\prime}}} \Pr\Big( \big\{x_k^k \,|\, S \setminus R(\pi_{k-1}) \big\} \ \big|\ \pi_{m-1}' \oplus x_{k-1}^k\Big) 
	\cdot 
	\Pr \Big( \pi_{m-1}' \oplus x_{k-1}^k \,\big|\, \big\{\pi_{k-1}|S \big\} \Big) \\
	&= c  \sum_{m=k-1}^{n-M+k-1} \sum_{\substack{\pi_{m-1}' \in \Sigma_{m-1}^{\prime}}} \Pr \Big( \pi_{m-1}' \oplus x_{k-1}^k \,\big|\, \big\{\pi_{k-1}|S \big\} \Big) \\
	&= c.
	\end{align*}
	
	We lay out the reasons for each equality below:
	\begin{enumerate}[label= (Equality \alph*):,leftmargin=2.3 cm]
	\baselineskip=17pt
	\item[($ 1^{st} $ equality:)] Rule of total probability.
	\item[($ 2^{nd} $ equality:)] Given $ m \in \{k-1, k, \ldots, k-1 + n - M\} $ and $ \pi_{m-1}' \in \Sigma'_{m-1} $, let 
	$$ c:= \Pr\Big( \big\{x_k^k \,|\, S \setminus R(\pi_{k-1}) \big\} \ \big|\ \pi_{m-1}' \oplus x_{k-1}^k\Big).$$ 
	be the probability that a (randomly drawn) participant chooses item $x_k^k$ out of the display set $S \backslash R(\pi_{k-1})$ given that he (she) ranks $\pi_{m-1}'$ as his (her) top-$ (m-1) $ preferred items and item $x_{k-1}^k$ as his (her) $m^{th}$ preferred one. By Lemma \ref{lma:oam_closed_form}, we have 
	\begin{align}\label{eq:second_part_RHS}
	c =
	\begin{cases}
	\tfrac{q^{|S|-k-\ssum_{x \in S \setminus R(\pi_{k-1})} \mathbb{I} \{x_k^k < x < x_{k-1}^k\}}}{1+q+\ldots+q^{|S|-k}} & \text{ if }   x_{k-1}^k > x_{k}^k , \\
	\tfrac{q^{\ssum_{x \in S \setminus R(\pi_{k-1})} \mathbb{I} \{x_{k-1}^k < x < x_k^k\}}}{1+q+\ldots+q^{|S|-k}} & \text{  if }  x_{k-1}^k < x_{k}^k.
	\end{cases}
	\end{align}
	Note that it is a constant independent of both $m$ and $\pi_{m-1}'$ (given that $ \pi_{k} $ and display set $ S $ are fixed). 
	
	\item [($ 3^{rd} $ equality:)] Note that $$\Pr \left( \pi_{m-1}' \oplus x_{k-1}^k \,\big|\, \left\{\pi_{k-1}|S \right\} \right) = \tfrac{\Pr \left( \pi_{m-1}' \oplus x_{k-1}^k  \text{ and } \left\{\pi_{k-1}|S \right\} \right)}{\Pr \left(\left\{\pi_{k-1}|S \right\} \right)}= \tfrac{\Pr \left( \pi_{m-1}' \oplus x_{k-1}^k \right)}{\Pr \left(\left\{\pi_{k-1}|S \right\}\right)}.$$
	Therefore, 
	\begin{align*}
	&\sum_{m=k-1}^{n-M+k-1} \sum_{\substack{\pi_{m-1}' \in \Sigma_{m-1}^{\prime}}} \Pr \Big( \pi_{m-1}' \oplus x_{k-1}^k \,\big|\, \big\{\pi_{k-1}|S \big\} \Big) \\
	= &\sum_{m=k-1}^{n-M+k-1} \sum_{\substack{\pi_{m-1}' \in \Sigma_{m-1}^{\prime}}} \tfrac{\Pr \left( \pi_{m-1}' \oplus x_{k-1}^k \right)}{\Pr \left(\left\{\pi_{k-1}|S \right\}\right)} \\
	= &\frac{1}{\Pr \left(\left\{\pi_{k-1}|S \right\}\right)} \sum_{m=k-1}^{n-M+k-1} \sum_{\substack{\pi_{m-1}' \in \Sigma_{m-1}^{\prime}}} \Pr \left( \pi_{m-1}' \oplus x_{k-1}^k \right) \\
	= &\frac{\Pr \left(\left\{\pi_{k-1}|S \right\}\right)}{\Pr \left(\left\{\pi_{k-1}|S \right\}\right)} = 1.
	\end{align*}
	\end{enumerate}

    Combining \eqref{eq:first_part_RHS} and \eqref{eq:second_part_RHS}, we have
    \begin{align*}
    \Pr(\{\pi_k|S\}) = &c  q^{d_S(\pi_{k-1})+L_S(\pi_{k-1})} \cdot \frac{\psi (|S|-k+1,q)}{\psi (|S|,q)} \\
    = &q^{d_S(\pi_{k})+L_S(\pi_{k})} \cdot \tfrac{\psi (|S|-k,q)}{\psi (|S|,q)}.
    \end{align*}
     Hence our statement holds for $k = K$, too, thus finishing the proof.
\end{proofref}

\vspace{0.4 cm}

\begin{proofref}{\cref{thm:ranked_choice_estimation}}
Suppose $\pi_{k} = (x_1, \ldots, x_k)$ is an arbitrary top-$k$ ranking. Recall from \cref{thm:multiple choice probability} that $d_{S} (\cdot)$ and $L_{S} (\cdot)$ are discrepancy measures with respect to the identity ranking. Through the relabeling argument explained in \cref{sec:RMI model}, we may generalize the discrepancy measures $d_{S} (\cdot)$ and $L_{S} (\cdot)$ to those with respect to an arbitrary (complete) ranking $ \pi $. Formally, let  
$$d^{\pi}_{S} := \sum_{h=1}^{k-1} (|S|-h) \cdot \mathbb{I}\{\pi^{-1}(x_h) >\pi^{-1}(x_{h+1})\}$$ and $$L^{\pi}_{S} := \sum_{j \in S \setminus \{x_1,\ldots,x_{k-1}\}} \mathbb{I}\{ \pi^{-1}(x_k) > \pi^{-1}(j)\},$$
respectively. Let the voting history be $H_{T}=\left(S_{1}, \pi_{k}^{1}, \ldots, S_{T}, \pi_{k}^{T}\right)$, where $\pi_{k}^{t} = (x_1^t, \ldots, x_k^t)$ and it's the $t^{th}$ participant's top-$k$ preferred items. Invoking \cref{thm:multiple choice probability}, the log likelihood for the RMJ-based ranking model parameter $(\pi, q)$  is
	\begin{align}
	\sum_{t=1}^{T} \log Pr^{\pi}(\pi_{k}^{t} \mid S_t) 
	&=\sum_{t=1}^{T} \left[ \log \frac{\psi (|S_t|-k,q)}{\psi (|S_t|,q)} +  \log q \cdot \big(d^{\pi}_{S_t}(\pi_{k}^{t})+L^{\pi}_{S_{t}}(\pi_{k}^{t})\big)  \right] \notag \\
	&=\sum_{t=1}^{T}  \sum_{i=|S_t|-k+1}^{|S_t|} \log \frac{1-q}{1-q^i} +  \log q  \cdot \underbrace{\left( \sum_{t=1}^{T}  d^{\pi}_{S_t}(\pi_{k}^{t})+L^{\pi}_{S_{t}}(\pi_{k}^{t}) \right)}_{*}.
	\end{align}
	As seen from above, the log likelihood is an affine transformation of the term $(*)$. In other words, there exist constants $a, b$, both independent of $\pi$, such that the log likelihood value can be written as $a + b \times (*)$. Now let us take a closer look at the term $(*)$.
	\begin{equation*}
	\begin{split}
	(*) = &\sum_{t=1}^{T}  d^{\pi}_{S_t}(\pi_{k}^{t})+L^{\pi}_{S_{t}}(\pi_{k}^{t}) \\
	    = &\sum_{t=1}^{T} \bigg(\sum_{h=1}^{k-1} (|S_t|-h) \cdot \mathbb{I}\{\pi^{-1}(x_h^t) >\pi^{-1}(x_{h+1}^t)\} + \sum_{j \in S_t \setminus \{x_1^t,\ldots,x_{k-1}^t\}} \mathbb{I}\{ \pi^{-1}(x_k^t) > \pi^{-1}(j)\} \bigg) \\
	    = &\sum_{t=1}^{T} \bigg(\sum_{h=1}^{k-1} (|S_t|-h) \cdot \sum_{(i, j): i \neq j} \mathbb{I}\{ \pi^{-1}(i) > \pi^{-1}(j)\} \cdot \mathbb{I}\{ x_h^t = i, x_{h+1}^t = j\} + \\ 
    &   \sum_{(i, j): i \neq j} \mathbb{I}\{ \pi^{-1}(i) > \pi^{-1}(j)\} \cdot \mathbb{I}\{ x_k^t =      i\} \cdot \mathbb{I} \{j \in S_t \setminus R(\pi_k^t) \} \bigg) \\
	= &\sum_{(i, j): i \neq j}   \mathbb{I}\{ \pi^{-1}(i) > \pi^{-1}(j) \} \cdot \bigg(\sum_{t=1}^{T}  \bigg[ \sum_{h=1}^{k-1} (|S_t|-h) \cdot \mathbb{I}\{ x_h^t = i, x_{h+1}^t = j\} + \\ 
	&  \mathbb{I}\{ x_k^t = i\} \cdot \mathbb{I} \{j \in S_t \setminus R(\pi_k^t) \}\bigg]   \bigg) . 
	\end{split}
	\end{equation*} 
	Comparing the form above and Proposition 3 of \cite{feng2021robust}, we conclude that to obtain the MLE for the central ranking $ \pi^\ast $, it suffices to solve the integer program \eqref{eqn:choiceAgg} with the generalized definition of  $w_{ij}$ below:
	$$
	w_{ij} := \sum_{t=1}^{T}  \bigg[ \mathbb{I}\{ x_k^t = i\} \cdot \mathbb{I} \{j \in S_t \setminus R(\pi_k^t) \} +  \sum_{h=1}^{k-1} (|S_t|-h) \cdot \mathbb{I}\{ x_h^t = i, x_{h+1}^t = j\}     \bigg].  
	$$
	
	That finishes the proof.
\end{proofref}  

\section{The Expectation-Maximization (EM) for for (Ranked) Choices under the RMJ-based Ranking Model}

In this section, we describe the standard expectation maximization (EM) algorithm used to fit a mixture of $M$ RMJ-based ranking models from (ranked) choices. We break the discussion into two steps: When $ k = 1 $, the data consists of single choices from (different) display sets. Every cluster follows a choice model described in \cref{thm:choice probability}, which is equivalent to the Ordinal Attraction model (OAM). When $ k \geq 1 $, the data consists of top-$ k $ ranked lists, and every cluster follows a ranked choice model described in \cref{thm:multiple choice probability}.

\subsection{Single choices ($ k=1 $)}

In this subsection, we describe the standard expectation maximization (EM) algorithm used to fit a mixture of $M$ OAMs to a given sample of choice data. This particular version is based on \cite{Antoine2021} and adapted for the OAM setting.

First, introduce the following notation.
\begin{itemize}
	\item $M$: number of clusters. 
	\item $Z_t:(c_{t1},...,c_{tM})$, $c_{tm} \in \{0,1\}$ indicates whether sample $t$ comes from cluster $m$.
	\item $\Theta = (\{p_m\},\{\pi_{m}\},\{\alpha_m\})$: $p_m$: mixture probability of cluster $m$, $\pi_m$ : central ranking of cluster $m$, $\alpha_m$: concentration parameter of cluster $m$.
	\item $OAM_m(x|S)$: choice probability of item $x$ given display set $S$ under cluster $m$.
\end{itemize}

To fit the mixture distribution, we would like to solve the following maximum likelihood problem:
$$
\max _{p, \pi, \alpha} \sum\limits_{m=1}^{M} \sum\limits_{t=1}^{T} c_{tm}\log p_m + \sum\limits_{m=1}^{M} \sum\limits_{t=1}^{T} c_{tm} \log [OAM_m(x_t|S_t)]
$$

The EM algorithm starts with an initial solution and iteratively obtains an improving solution until an appropriate stopping criterion is met. Suppose $\left\{\left(p_{m}, \pi_{m}, \alpha_{m}\right): 1 \leq m \leq M\right\}$ is the current solution. Then, an improving solution $\left\{\left(\hat{p}_{m}, \hat{\pi}_{m}, \hat{\alpha}_{m}\right): 1 \leq m \leq M\right\}$ is obtained as follows.

In the Expectation step, the algorithm computes "soft counts" $c_{t m}$, denoting the (posterior) probability that example $t$ was generated from mixture component $m$. This probability is computed as
$$ \hat{c}_{tm}=\frac{\hat{p}_m*\hat{OAM}_m(x_t|S_t)}{\sum_{j=1}^{M} \hat{p}_{j}*\hat{OAM}_{j}(x_t|S_t)}$$

Then, in the Maximization step, we first set $\hat{p}_m = \frac{1}{T}\sum\limits_{t=1}^{T} \hat{c}_{tm}$, and then solve a separate optimization problem for each segment $1 \leq m \leq M$ :
$$
\max _{\pi_{m}, \alpha_{m}}-\alpha_m \sum\limits_{t=1}^{T} \hat{{c}}_{t m} \ssum_{x \in S_t} \I{x \succ_{\pi_m} x_t} - \sum\limits_{t=1}^{T} \hat{{c}}_{t m} \log {\sum\limits_{j=0}^{|S_t|-1} e^{-\alpha_m j} }.
$$

We can obtain an optimal solution of the above problem by solving the following two problems:

\begin{align}
d_{m}^{*}&=\min _{\pi_{m}} \sum_{t=1}^{T} \hat{c}_{t m} \cdot \ssum_{x \in S_t} \I{x \succ_{\pi_m} x_t} \label{eqn:dm}\\
\hat{\alpha}_{m}&=\arg \min _{\alpha_{m}}\left\{\alpha_{m} \cdot d_{m}^{*}+\sum\limits_{t=1}^{T} \hat{c}_{t m} \log {\sum\limits_{j=0}^{|S_t|-1} e^{-\alpha_m j} }\right\} \label{eqn:alpham}
\end{align}
The problem in \eqref{eqn:dm} can be solved by choice aggregation through the integer program \eqref{eqn:choiceAgg} (with properly defined $ w $ scores). Also, the optimization problem in \eqref{eqn:alpham} is convex. 

Once we obtain the new solution, $\left\{\left(\hat{p}_{m}, \hat{\pi}_{m}, \hat{\alpha}_{m}\right): 1 \leq m \leq M\right\}$, the above process is repeated until the stopping criterion is met. Our stopping criterion is as follows:
\begin{enumerate}
    \item $\{\hat{\pi}_m: 1 \leq m \leq M \}$ don't change;
    \item $||\hat{\boldsymbol{p}}_{\text{current  round}}-\hat{\boldsymbol{p}}_{\text{previous round}}||_1 < 1e-3$
     or $||\hat{\boldsymbol{\alpha}}_{\text{current  round}}-\hat{\boldsymbol{\alpha}}_{\text{previous  round}}||_1 < 1e-3$, where $\boldsymbol{p}=(p_1,p_2,\ldots,p_M)$ and $\boldsymbol{\alpha} = (\alpha_1,\alpha_2,\ldots,\alpha_M).$
\end{enumerate}

Finally, since the EM algorithm is only guaranteed to converge to local optimality, we run (in parallel) 20 instances the EM algorithm with random initialization and select the parameters with the best log-likelihood value.

\subsection{The general setting of ranked choices ($ k\geq 1 $)}

In this subsection, we describe the standard expectation maximization (EM) algorithm used to fit a mixture of $M$ RMJ-based ranking models from a sample of top-$k$ ranked choice data. This particular version is based on \cite{Antoine2021} and adapted for the ranked choice setting.

First, introduce the following notation.
\begin{itemize}
    \item $k$: number of ranked items in each sample
	\item $M$: number of clusters. 
	\item $Z_t:(c_{t1},...,c_{tM})$, $c_{tm} \in \{0,1\}$ indicates whether sample $t$ comes from cluster $m$.
	\item $\Theta = (\{p_m\},\{\pi_{m}\},\{\alpha_m\})$: $p_m$: mixture probability of cluster $m$, $\pi_m$ : central ranking of cluster $m$, $\alpha_m$: concentration parameter of cluster $m$.
	\item $RCM_m(\pi_k|S)$: ranked choice probability of $\pi_k$ given display set $S$ under cluster $m$.
\end{itemize}

To fit the mixture distribution, we would like to solve the following maximum likelihood problem:
$$
\max _{p, \pi, \alpha} \sum\limits_{m=1}^{M} \sum\limits_{t=1}^{T} c_{tm}\log p_m + \sum\limits_{m=1}^{M} \sum\limits_{t=1}^{T} c_{tm} \log [RCM_m(\pi_k^t|S_t)]
$$

The EM algorithm starts with an initial solution and iteratively obtains an improving solution until an appropriate stopping criterion is met. Suppose $\left\{\left(p_{m}, \pi_{m}, \alpha_{m}\right): 1 \leq m \leq M\right\}$ is the current solution. Then, an improving solution $\left\{\left(\hat{p}_{m}, \hat{\pi}_{m}, \hat{\alpha}_{m}\right): 1 \leq m \leq M\right\}$ is obtained as follows.

In the Expectation step, the algorithm computes "soft counts" $c_{t m}$, denoting the (posterior) probability that example $t$ was generated from mixture component $m$. This probability is computed as
$$ \hat{c}_{tm}=\frac{\hat{p}_m*\hat{RCM}_m(\pi_k^t|S_t)}{\sum_{j=1}^{M} \hat{p}_{j}*\hat{RCM}_{j}(\pi_k^t|S_t)}$$

Then, in the Maximization step, we first set $\hat{p}_m = \frac{1}{T}\sum\limits_{t=1}^{T} \hat{c}_{tm}$, and then solve a separate optimization problem for each segment $1 \leq m \leq M$ :
$$
\max _{\pi_{m}, \alpha_{m}}-\alpha_m \sum\limits_{t=1}^{T} \hat{{c}}_{t m} (d_{S_t}(\pi_k^t)+L_{S_t}(\pi_k^t)) - \sum\limits_{t=1}^{T} \hat{{c}}_{t m} \sum\limits_{i=n-k+1}^{n} \log {\sum\limits_{j=0}^{i-1} e^{-\alpha_m j}}
$$

We can obtain an optimal solution of the above problem by solving the following two problems:

\begin{align}
d_{m}^{*}&=\min _{\pi_{m}} \sum\limits_{t=1}^{T} \hat{{c}}_{t m} (d_{S_t}(\pi_k^t)+L_{S_t}(\pi_k^t)) \label{eqn:dm_topk}\\
\hat{\alpha}_{m}&=\arg \min _{\alpha_{m}} \left\{\alpha_{m} \cdot d_{m}^{*}+\sum\limits_{t=1}^{T} \hat{{c}}_{t m} \sum\limits_{i=n-k+1}^{n} \log {\sum\limits_{j=0}^{i-1} e^{-\alpha_m j}} \right\} \label{eqn:alpham_topk}
\end{align}
The problem in \eqref{eqn:dm_topk} can be solved by choice aggregation through the integer program \eqref{eqn:choiceAgg} (with $ w $ defined in Theorem \ref{thm:ranked_choice_estimation}). Also, the optimization problem in \eqref{eqn:alpham_topk} is convex. 

Once we obtain the new solution, $\left\{\left(\hat{p}_{m}, \hat{\pi}_{m}, \hat{\alpha}_{m}\right): 1 \leq m \leq M\right\}$, the above process is repeated until until the stopping criterion is met. Our stopping criterion is as follows:
\begin{enumerate}
    \item $\{\hat{\pi}_m: 1 \leq m \leq M \}$ don't change;
    \item $||\hat{\boldsymbol{p}}_{\text{current  round}}-\hat{\boldsymbol{p}}_{\text{previous round}}||_1 < 1e-3$
     or $||\hat{\boldsymbol{\alpha}}_{\text{current round}}-\hat{\boldsymbol{\alpha}}_{\text{last  round}}||_1 < 1e-3$, where $\boldsymbol{p}=(p_1,p_2,\ldots,p_M)$ and $\boldsymbol{\alpha} = (\alpha_1,\alpha_2,\ldots,\alpha_M).$
\end{enumerate}

Finally, since the EM algorithm is only guaranteed to converge to local optimality, we run (in parallel) 20 instances the EM algorithm with random initialization and select the parameters with the best log-likelihood value.

\section{More Experiments} \label{more experiments}

\textbf{Experiment 1.1: More analysis on the results of experiment 1.} In experiment 1 of section \ref{numerical}, We randomly split the data into 4000 preference rankings in training and 1000 preference rankings in testing. When we train mixed-MNL on the full set, the mixed-MNL with only one cluster can already perfectly explain the choice data. That is, the predicted choice probabilities are exactly equal to the empirical choice probabilities (aka the market shares). Hence, the learned parameters of mixed-MNL with a smaller number of clusters are also optimal solutions to the mixed-MNL with a larger number of clusters. Specifically, the optimal parameters of $i$-cluster MNL , $i \in \{1,2,\ldots,k\}$ are all optimal to $k$-cluster MNL. In other words, mixed-MNL with multiple clusters has many optimal solutions. 

\begin{figure}[htbp] 
	\centering
	\caption{Train on the full display set only (Each row is on a random data splitting.) }
	\includegraphics[width=1.0 \textwidth]{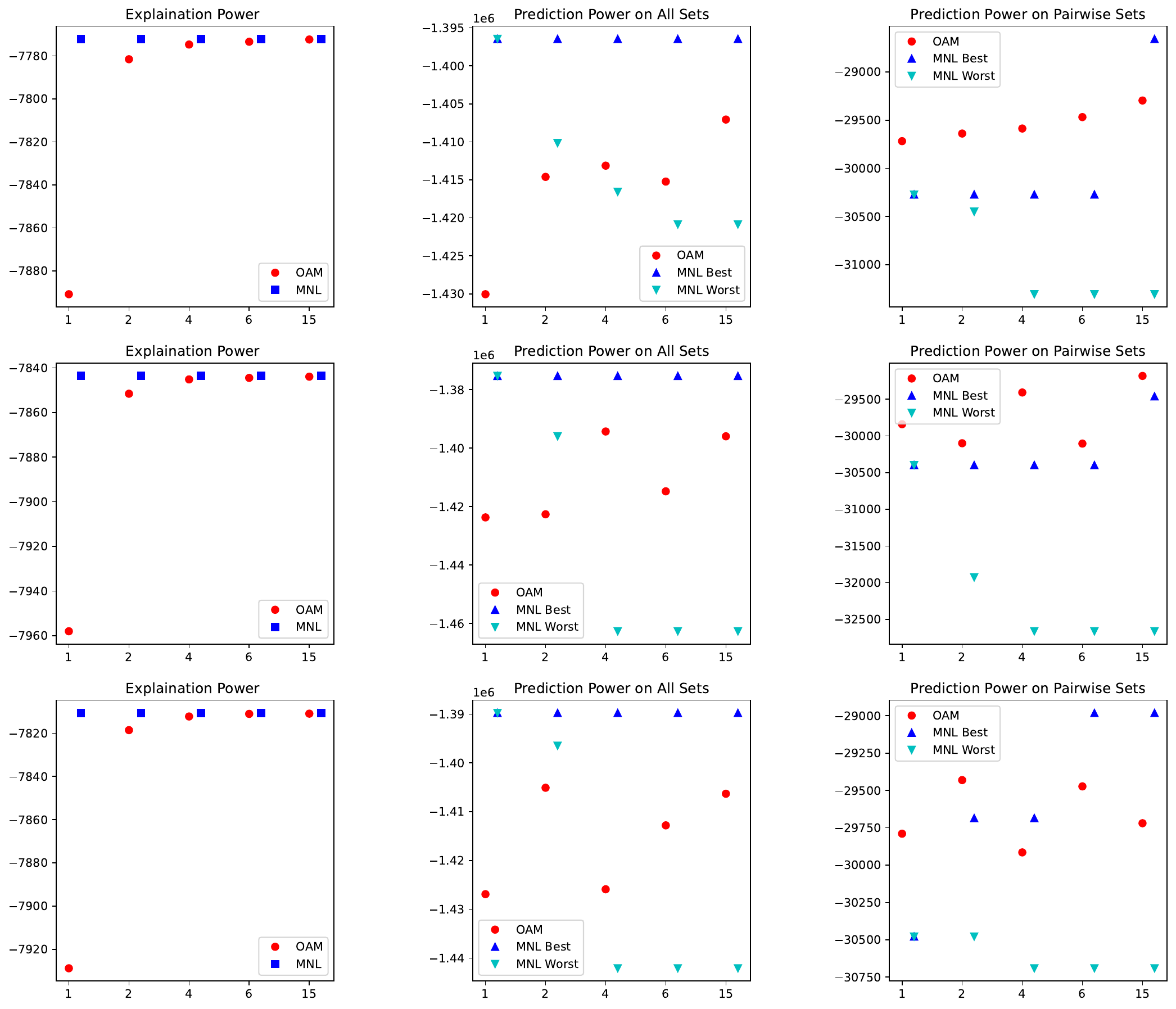}
	\ \\
	{\footnotesize \sf In each panel, the x-axis represents the number of clusters in the mixture model, and the y-axis represents the log-likelihood metric.}
	\label{fig:mnl_range}
\end{figure}

These optimal parameters can predict the same choice probabilities on the full set but may predict differently on other sets. Given the number of clusters, we record the best log likelihood value and worst log likelihood value from the multiple optimal parameters.
We repeat the random splitting 3 times and show these 3 results in the \cref{fig:mnl_range}.

In these three figures, fix the number of clusters, the prediction power of mixed-MNL may span wide ranges between the best record and the worst record.\footnote{Since we only try out a limited number of optimal solutions, the actual range could be much wider.} However, the prediction power of OAM are in or beyond the range across different number of clusters under two test setups.

\textbf{Experiment 1.2: Repeatedly shuffle the data.} Now, we repeat the data splitting 80 times to obtain multiple performance values. 

We use $\bar{ll}$ to denote the mean of log-likelihood values from the 80 times data splitting and $std(ll)$ to denote the standard deviation of log-likelihood values. The confidence interval (CI) is defined as $[\bar{ll}-1.96 \cdot \frac{std(ll)}{\sqrt{SS}}, \bar{ll}+1.96 \cdot  \frac{std(ll)}{\sqrt{SS}}]$, where the sample size $SS=80$.

\begin{figure}[htbp] 
	\centering
	\caption{Train on full set cross shuffle}
	\includegraphics[width=1.0 \textwidth]{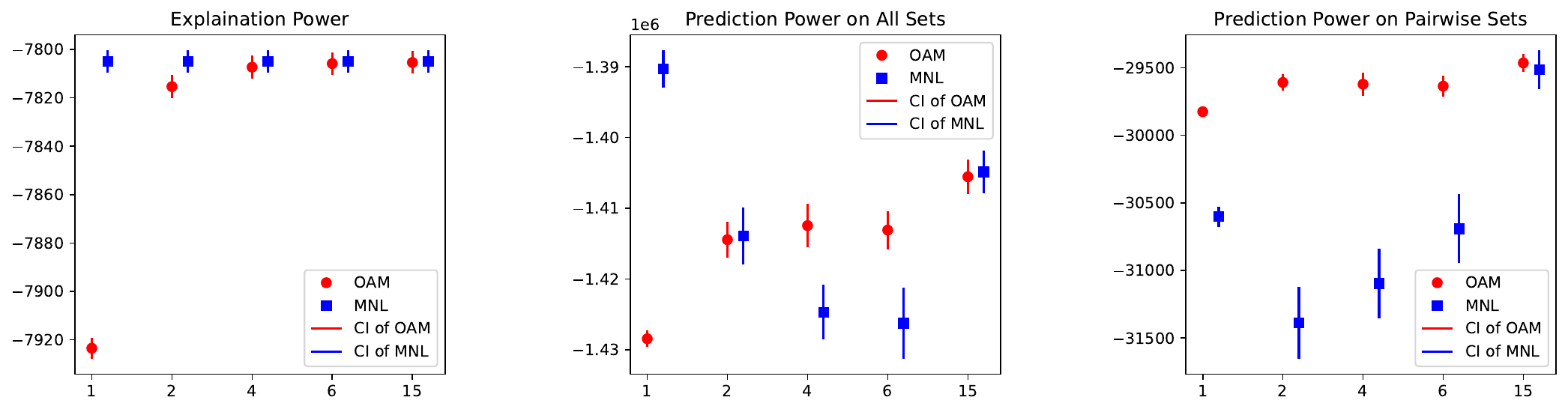}
	\label{fig:full_cross}
\end{figure}

\begin{figure}[htbp] 
	\centering
	\caption{Train on three random display sets cross shuffle}
	\includegraphics[width=1.0 \textwidth]{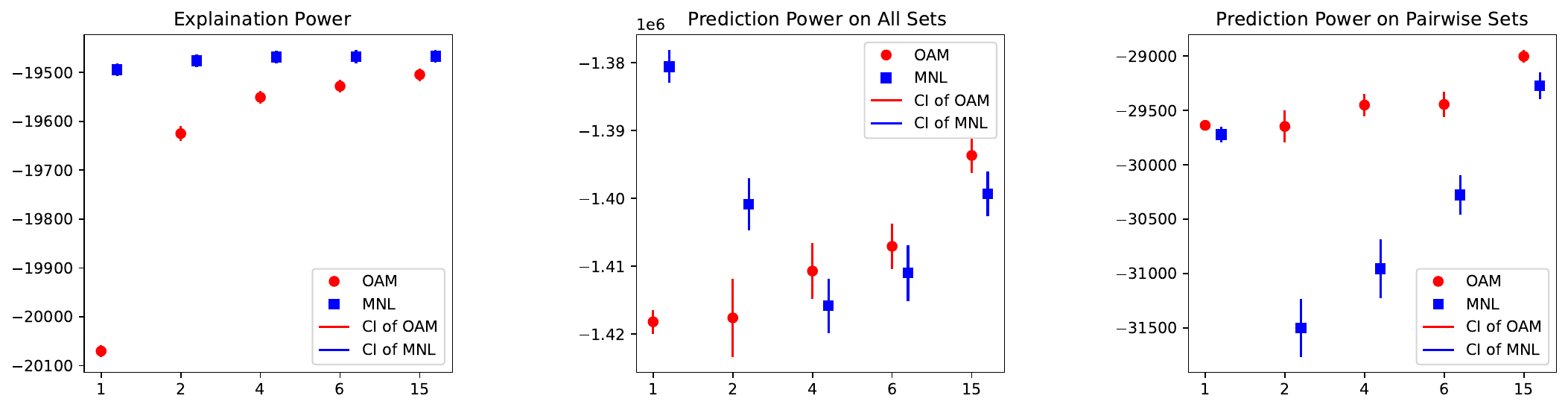}
	\label{fig:fix_cross}
\end{figure}

The result of training on the full set is in \cref{fig:full_cross}. Given the collection of random sets in experiment 1 of section \ref{numerical}, the result is in \cref{fig:fix_cross}. From these two figures, the prediction power of OAM increases with the number of clusters increases. Hence, OAM seems to suffer less from overfitting.

\textbf{Experiment 2.1: Compare with Borda Count and Simple Count.} Since classical Borda Count could only be applied on the full set, we adapt it as follows: If a display set contains $M$ items and there is a top-$k$ ranked choice, the ranked items in the ranked choice get $(M-1),(M-2),\ldots,(M-k)$ scores. We also consider an even simpler variation of the Borda count where the scores are unweighted. Because of the simplicity of this variation, we refer to it as Simple Count. 

First, we compare three methods on the setups used in experiment 2 of section \ref{numerical}, where all sets with sizes larger or equal to $k$ will be displayed when the feedback is top-$k$ ranked choice. The learned rankings under top-$1$, top-$2$, top-$3$ feedback structure are shown in \cref{tab:ranking_balance}.
\begin{table}[htbp]
	\centering
	\begin{tabular}{|c|c|c|c|}
		\hline
		& OAM                    & Borda Count            & Simple Count           \\ \hline
		Top-3 & (8,5,6,3,2,1,4,9,7,10) & (8,3,6,1,2,5,4,9,7,10) & (8,3,1,6,2,5,4,9,7,10) \\ \hline
		Top-2 & (8,5,6,3,2,1,4,9,7,10) & (8,3,5,6,1,2,4,9,7,10) & (8,3,6,1,5,2,4,9,7,10) \\ \hline
		Top-1 & (8,5,6,3,2,1,4,9,7,10) & (8,5,6,3,2,1,4,9,7,10) & (8,5,3,6,2,1,4,9,7,10) \\ \hline
	\end{tabular}
	\vskip 0.1 cm
	\caption{Learned rankings under different feedback structure when display sets are balanced.}
	\label{tab:ranking_balance}
\end{table}

To measure the discrepancy among the estimated rankings from top-1, top-2, and top-3 feedback, we use the average pairwise Kendall's Tau distance among the estimated rankings. The results are  $0,4,$ and $4$, respectively, for OAM, Borda Count, and Simple Count, respectively. As a result, we can see that OAM is the most stable one.

Furthermore, and perhaps more importantly, Borda Count and Simple Count become particularly ineffective when the display sets are unbalanced because those simple metrics fail to incorporate the display set information. For example, if the learner disproportionally displays the least preferred items (items 7,9,10 in our data), those items will be ranked high under Borda/simple counts because only the ``counts" matter. In comparison, the RMJ-based method judiciously adjusts the count of an item by the display set history.

\begin{table}[htbp]
	\centering
	\begin{tabular}{|c|c|c|c|}
		\hline
		& OAM                    & Borda Count          & Simple Count           \\ \hline
		Top-3 & (8,5,3,2,6,1,4,9,7,10) & (8,3,5,6,9,2,1,7,4,10) & (9,7,10,8,3,5,6,2,1,4) \\ \hline
		Top-2 & (8,5,6,3,2,1,4,9,7,10) & (8,5,9,6,3,2,1,7,4,10) & (9,7,8,10,5,6,3,2,1,4) \\ \hline
		Top-1 & (8,5,2,6,1,3,4,9,7,10) & (8,5,9,7,2,6,1,3,4,10) & (9,7,8,5,2,6,10,1,3,4) \\ \hline
	\end{tabular}
	\vskip 0.1 cm
	\caption{Learned rankings under different feedback structures when display sets are unbalanced.}
	\label{tab:ranking_unbalance}
\end{table}

In \cref{tab:ranking_unbalance}, we show the learned rankings under top-$1$, top-$2$, top-$3$ feedback structure when only the full set $[10]$ and $\{7,9,10\}$ are displayed. The average pairwise Kendall's Tau distance values of these three methods are $2.67,7.33$ and $6$. OAM is still the most stable one under this unbalanced setting. In addition, from \cref{tab:ranking_unbalance}, OAM still ranks item $7,9,10$ at the bottom, while Borda Count puts items $7,9$ at some high positions and Simple Count put items $7,9,10$ at very top positions. 

Let us summarize our findings in Experiment 2.1 and the advantage of OAM compared to different variants of the Borda count. First, under both balanced and unbalanced settings, OAM is the most stable method. Second, Borda Count and Simple Count will learn misleading ranking when the display sets are unbalanced, i.e., some items are displayed much more times than other items, while OAM won't be affected by this unbalance. Finally, it is worth pointing out that Borda Count and Simple Count are only about the central ranking, while OAM is also about making (probabilistic) predictions on ranked choices. 

\textbf{Discussion: further comments on the numerical studies.}  We believe that our numerical experiments have demonstrated promising evidence that the RMJ-based ranking model can be used to capture people's ranked choices from varied display sets. As such, it adds to the toolbox for our motivating application in preference learning in contexts such as crowdsourcing, marketing research, and survey designs. Since our model is a generalization of choice modeling (i.e., every participant makes a single choice out of a subset of items), we believe it can find potential applications in other domains such as online/brick-and-mortar retailing. Of course, more comprehensive numerical experiments are called for to better evaluate its potential, which we leave for future research. For example, we could use data sets that specialize in the retail setting. In addition, it would be interesting to see how the current method handles the non-purchase option since it may well be the overwhelmingly most frequent response in the data. Finally, it would be beneficial to explore how the current method can be extended to incorporate customer covariates. 

\textbf{Experiment 4: Experiment on the YOOCHOOSE dataset for e-commerce.} We wish to conclude this section by showing some initial sign of success in the e-commerce context. We compare our model with MNL on the YOOCHOOSE dataset of the RECSYS 2015 challenge, which contains six months of user activities for a large European e-commerce business \cite{ben2015recsys}.

Here is a brief description of the data set and the data cleaning rule. In this data set, click and purchase data are provided at the session level. In every session, we can observe the collection of products that the customer clicks and the collection of products that the customer buys. We use the click data to form the ``display sets" and purchase data to form the ``choices" in our paper.\footnote{We find that the no-purchase option is overwhelmingly frequent in the data. To deal with this issue and compare the customer preferences over the real items, we filter out the non-purchase and single-click sessions.} We conduct our experiment on category 12. In particular, we focus on the top 111 items, which consist of over $95\%$ click data. If there are purchases of multiple products in one session, we randomly choose one item as the customer's choice. If the customer buys more than one same item in a single session, we also treat it as a single choice. In the end, there are 76 items and 274 choice data points after data cleaning. 

Here is a brief description of how to randomly shuffle and split the data into training and testing sets. To ensure that all items appear in the training set, we keep 53 choice data in the training set and randomly split the other 221 choice data into the training and testing set. Finally, there are 220 choice data in the training set and 54 choice data in testing. We randomly repeat the data splitting 100 times. On average, there are 124.75 unique display sets in each training set and 2.72 items in a display set. Note that this is an extremely sparse collection, considering the power set of 76 items consists of $ 2^{76} \approx 7.56 \times 10^{22}$ elements.

\begin{figure}[htbp] 
	\centering
	\caption{Results of 100 times data splitting on the YOUCHOOSE dataset}
	\includegraphics[width=0.8 \textwidth]{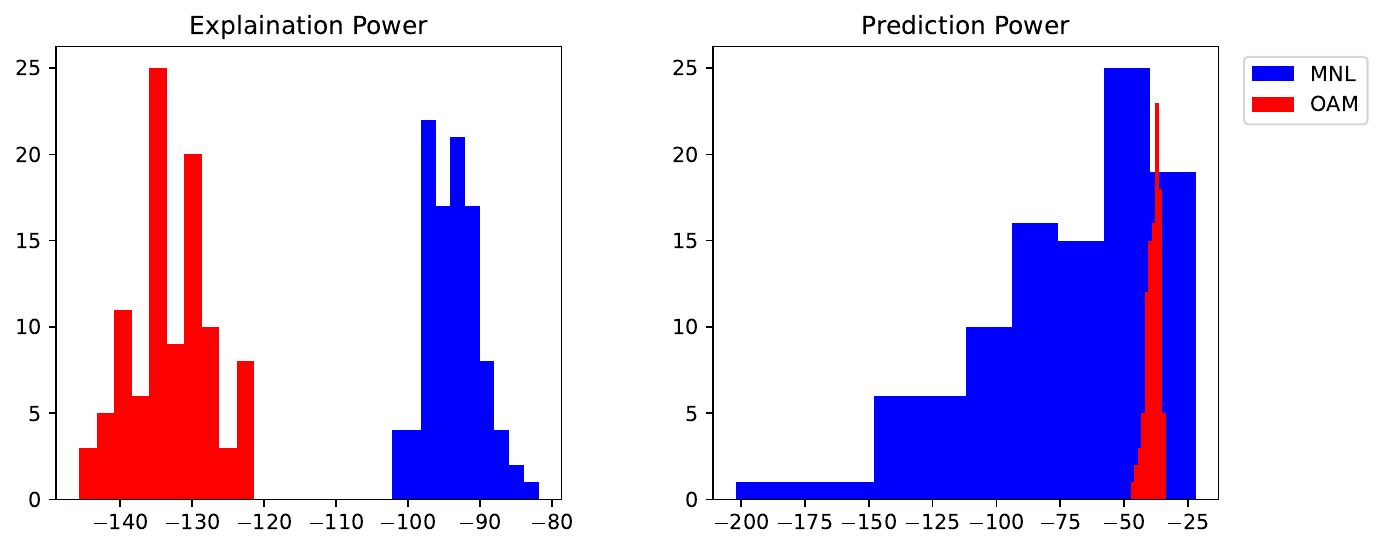}
	\ \\
	{\tiny \sf In each panel, the x-axis represents the log-likelihood metric and the y-axis represents the frequency. }
	\label{fig:yoochoose}
\end{figure}

In this experiment, we use one cluster OAM and MNL. The results are shown in \cref{fig:yoochoose}. From this figure, MNL achieves better explanation power, but OAM performs better prediction power on average. This is a sign that MNL tends to overfit in this data set where the coverage of the display set is limited. In comparison, OAM demonstrates better generalization ability.

%

\end{document}